\documentclass[twoside,11pt]{article}
\usepackage{xcolor}
\usepackage{blindtext}
\usepackage{tabularx} 
\usepackage{amsmath}
\usepackage[utf8]{inputenc}
\usepackage[most]{tcolorbox}
\tcbuselibrary{breakable}
\usepackage{multirow}

\usepackage[abbrvbib, preprint]{jmlr2e}

\usepackage{jmlr2e}

\usepackage{lastpage}
\jmlrheading{xx}{2025}{1-\pageref{LastPage}}{2/25; Revised xx/25}{xx/25}{21-0000}{Javier Coronado-Blázquez}

\ShortHeadings{LLMs as random number generators}{Coronado-Blázquez}
\firstpageno{1}

\begin{document}

\title{Deterministic or probabilistic? The psychology of LLMs as random number generators}

\author{\name Javier Coronado-Blázquez \email j.coronado.blazquez@gmail.com \\
       \addr Telefónica Tech, AI \& Data Unit\\
       Madrid, 28050, Spain}

\editor{My editor}

\maketitle

\begin{abstract}
Large Language Models (LLMs) have transformed text generation through inherently probabilistic context-aware mechanisms, mimicking human natural language. In this paper, we systematically investigate the performance of various LLMs when generating random numbers, considering diverse configurations such as different model architectures, numerical ranges, temperature, and prompt languages. Our results reveal that, despite their stochastic transformers-based architecture, these models often exhibit deterministic responses when prompted for random numerical outputs. In particular, we find significant differences when changing the model, as well as the prompt language, attributing this phenomenon to biases deeply embedded within the training data. Models such as DeepSeek--R1 can shed some light on the internal reasoning process of LLMs, despite arriving to similar results. These biases induce predictable patterns that undermine genuine randomness, as LLMs are nothing but reproducing our own human cognitive biases.
\end{abstract}


\section{Introduction}
\label{sec:intro}

Large Language Models (LLMs) have revolutionized natural language processing by generating human-like text through advanced probabilistic mechanisms. Based on transformer architectures \cite{vaswani2023attentionneed} and trained on vast corpora of text, these models learn to predict the next token in a sequence, effectively capturing intricate statistical patterns inherent in human language. Although LLMs are inherently stochastic, recent observations have revealed a curious phenomenon: when tasked with generating a single random number --a seemingly trivial exercise in randomness-- they often produce deterministic outputs. This counterintuitive behavior raises important questions about the interplay between a model’s probabilistic design and the biases ingrained in its training data.

LLMs are deep neural networks that leverage the transformer architecture to perform a wide range of natural language tasks. Transformers use self-attention mechanisms to model long-range dependencies in text, allowing the model to assign a probability distribution over possible next tokens based on context. Training involves maximizing the likelihood of observed sequences, which results in a model that can generate text by sampling from its learned distribution. In theory, such a mechanism should naturally yield variable and random outputs when the model is allowed to sample freely. However, the actual behavior of these models often deviates from this ideal, especially in tasks that require pure randomness.

The probabilistic nature of LLMs is central to their function. When generating text, each output token is sampled from a distribution conditioned on prior tokens, leading to variability and creativity. This stochastic process is expected to extend to all tasks, including the generation of a single random number. Yet, numbers are not understood as such by LLMs, but rather as tokens, attending to their characters and not their mathematical meaning. This is, a number such as ``2'' has no further meaning for a LLM than ``3'', ``+'' or the word ``horse'' -- they are just tokens (either singular or a collection) with corresponding vector(s) in the latent space of the embedding model.

In an ideal setting, requesting a random number from an LLM should yield outputs that are uniformly distributed over the specified range. Yet, as noted in recent discussions and blog posts\footnote{See \href{https://medium.com/@kmaurinjones/evaluating-randomness-in-generative-ai-large-language-models-099f747b28e2}{``Evaluating Randomness in Generative AI \& Large Language Models"}}, many models exhibit a pronounced bias toward particular outputs when tasked with generating randomness. This observation suggests that the randomness encoded within the LLMs' sampling procedures may be compromised by factors beyond the mere sampling algorithm.

While LLMs are designed to generate outputs based on probabilistic principles, they are ultimately trained on human-generated text, which is replete with patterns, conventions, and biases. These training datasets often include overrepresented sequences and stylistic regularities that can skew the learned probability distributions. Consequently, when an LLM is prompted to generate a random number, it may default to outputs that reflect these ingrained patterns. This phenomenon aligns with the ``stochastic parrot" critique \cite{10.1145/3442188.3445922}, where models are seen as reproducing the statistical regularities of their training data without true understanding.

The issue of stochasticity in LLM outputs has garnered increasing attention in both academic and informal settings. A recent study \cite{vankoevering2024randomrandomevaluatingrandomness} systematically analyzed the randomness of outputs from several popular models and found that certain systems deviate markedly from expected behavior.

Additionally, the choice of sampling parameters such as the temperature, top‑k, or top-p play a significant role in balancing randomness and determinism. Lower temperature values, for instance, concentrate the probability mass and should lead to more deterministic outputs. Even when these parameters are adjusted to encourage variability, many LLMs still tend to output similar ``random" numbers repeatedly, hinting that the bias is deeply embedded in the model’s internal representations and the nature of its training data.

The deterministic tendencies observed in random number generation also have broader implications. In applications where true randomness is essential --for example, in cryptographic protocols, statistical sampling, or even in certain simulation tasks-- the inability of LLMs to generate uniformly random outputs could lead to significant vulnerabilities or performance issues. Understanding these limitations is therefore not only of theoretical interest but also of practical importance.

In this paper, we conduct a systematic investigation into the capability of LLMs to function as random number generators. We explore a range of configurations including different model architectures, numerical ranges, and --for the first time-- languages to assess how these factors influence the randomness of the outputs. While other works' conclusions point towards inherited biases from the training data in answers, none has explored the influence of the prompt language to check whether there are significant differences depending on it.

This idiomatic tests have been widely explored for possible cultural and linguistic biases in natural language answers (see, e.g., \cite{neplenbroek2024mbbqdatasetcrosslingualcomparison, mihaylov2024elegantbridgemultilingualllms, Tao_2024}). The issue of random number generation in LLMs has been previously tackled by \cite{hopkins2023can}, but for entire sequences to study the uniformity of those. The authors find that LLMs do not always generate the expected distribution, breaking the uniformity assumption that is required in the prompt. In our case, we are interested in the LLM variability of a single number per call, and how can they reproduce human biases when offered such choice. 

One of the models evaluated in this work is DeepSeek--R1, which outputs not only the answer but the full reasoning chain-of-thought (CoT) to determine the final output. This offers a novel and rich view at the internal process of an LLM when prompted to generate a random number, as often we find a very extensive monologue with several changes of mind between, yet arriving most of the times to similar conclusions.

This study not only sheds light on the limitations of current LLMs as random number generators but also opens avenues for further research into mitigating data-induced determinism in probabilistic models. Addressing such issues is crucial for ensuring that LLMs can be reliably used in contexts where unpredictability and fairness are of paramount importance.

The remainder of the paper is organized as follows. Section \href{2}{\ref{sec:methodology}} describes our experimental setup, including the various configurations and methodology employed to probe the randomness of the models. Section 3 presents the results of our study, comparing the behavior of different models, highlighting key differences in output distributions and computing statistical tests. Finally, Section 5 concludes with a summary and directions for future research.

\section{Experimental setup and methodology}
\label{sec:methodology}

In order to evaluate the stochasticity capabilities of LLMs when tasked with generating a single random number, we conduct a systematic set of experiments covering multiple configurations. Specifically, we test the models alphabetically summarized in Table \ref{tab:models_summary}:

\begin{table}[!ht]
    \centering
    \begin{tabular}{c|c|c|c}
    Model name & Developer & Parameters & Access \\
    \hline
    \hline
    \bf DeepSeek-R1 & DeepSeek & 14B & Local \\
    \hline
    \bf Gemini 2.0 & Google & -- & API \\
    \hline
    \bf GPT-4o-mini & OpenAI & -- & API \\
    \hline
    \bf Llama 3.1 & Meta & 8B & Local \\
    \hline
    \bf Mistral & Mistral & 7B & Local \\
    \hline
    \bf Phi-4 & Microsoft & 14B & Local
    \end{tabular}
    \caption{Summary of the model pool evaluated in this work.}
    \label{tab:models_summary}
\end{table}

Due to computational restrictions, we do not use models with large number of parameters $(\gtrsim 20$B), although we do test Gemini 2.0 and GPT-4o-mini, with an unreported number of parameters but expected to be massive \cite{openai2024gpt4technicalreport, geminiteam2024gemini15unlockingmultimodal}. Likewise, we avoid Small Language Models (below $\sim5$B) as initial tests conducted with Llama 3.2--3B and Gemma--2B suggested these models had some difficulties to properly understand the task consistently.

Initially, we also included Perplexity's Sonar models, yet, as distilled from both DeepSeek and Llama families, we found their results to be very similar to those in preliminary tests. Being a pay-per-use model, we decided to exclude them from the model pool for resource efficiency sake. Additionally, we considered to use Qwen 2.5, but the 14B version of DeepSeek--R1 used in this study is distilled from the Qwen architecture \cite{deepseekai2025deepseekr1incentivizingreasoningcapability}, and decided to prioritize DeepSeek for its CoT reasoning.

For each model, the experiments are carried out in seven different languages: Chinese (CN), English (EN), French (FR), Hindi (IN), Japanese (JP), Russian (RU), and Spanish (ES). We select these languages based on two primary criteria: (i) they represent a broad spectrum of linguistic typologies with distinct grammatical and morphological features, as well as different cultural backgrounds, and (ii) they are among the most widely spoken languages globally and are well represented in the large-scale training corpora of modern LLMs.

The prompt is always the same: \texttt{Give me a random number between 1 and X. Please only return the number with no additional text}, where X is the upper limit defined in each of the three range configurations. We replicate it into the other 6 languages in their respective alphabets (e.g., Cyrillic for Russian). This prompt ensures that the task is well-defined, yet offers certain freedom (for example, we do not specify the number to be an integer). We deliberately do not add any prompt engineering such as ``make sure this number is truly random'' or ``avoid giving a deterministic answer'' to be able to spot possible biases in the generation process.

While these subtleties may seem irrelevant when prompting such a straightforward task as generating a single number, we find that language influences the distribution of the resulting samples, as most likely the model is unconsciously finding patterns in the corresponding language subset of the corpus. English comprises the vast majority of training corpora\footnote{English training tokens are reported to be 92.65\% for GPT-3.5 and 89.7\% for Llama 2 \cite{li2024languagerankermetricquantifying}.}, yet these languages are present in these models and can understand the task they are being prompted. In particular, both Llama and Gemini are models that excel as transfer for different languages, even for those with no representation in the training data in some cases \cite{akter2023indepthlookgeminislanguage, zhao2024llamaenglishempiricalstudy, guo2024benchmarkinglinguisticdiversitylarge}.

Furthermore, we evaluate the models under three distinct random number generation configurations: 1--5, 1--10 and 1--100 range, which are the typical ranges humans use when thinking about a number. Finally, we also perform six different temperature configurations: $T=\left[0.1,~0.3,~0.5,~0.8,~1.0,~2.0\right]$. This selection provides balance in the trade-off between granularity and computational cost. For each combination of model, language, and random number range, we performed 100 independent calls. The full setup encompasses:
\[
6 \text{ models} \times 7 \text{ languages} \times 3 \text{ ranges} \times 6 \text{ temperatures} \times 100 \text{ numbers} = \textbf{75600} \text{ calls}
\]

The experimental procedure is as follows: for each language and model, a prompt is constructed to request a random integer within the specified range. For the open-source models, we use Ollama integrated with Python, while for the proprietary models we use their respective APIs. The outputs are individually stored for further statistical analysis to determine the degree of randomness (or determinism) in the generated numbers. In subsequent sections, we detail the statistical metrics employed to evaluate the output distributions.

\section{Results and discussion}
\label{sec:results}

Results are stored in individual csv files for the analysis. By manually inspecting them, we find that, even if the prompt explicitly states that no further text is generated, some models often generate extra output, such as the examples shown in Table \ref{tab:phi4_extra}:

\begin{table}[!ht]
    \centering
    \begin{tabularx}{\textwidth}{X|c} 
        \bf Additional output & \bf Model \\
        \hline
        \hline
        (Note: As an AI model, I can't actually generate random numbers in real-time. The number provided is a placeholder for demonstration purposes.) & Phi-4 \\ \hline
        (Note: The number is randomly generated and will differ each time you ask for one.) & Phi-4  \\ \hline
        (Note: As an AI language model, I cannot generate true randomness. The number provided here is just a random choice within the specified range.) & Phi-4  \\ \hline
        (Note: Each request for a random number will generate a different result.) & Phi-4 \\
        \hline
        I wanted to reply to this answer as 47, but I needed to write a program to generate random numbers according to the prerequisites. To achieve this, you need to specify a programming language and library [...] & Mistral \\
        \hline
        Note: This number was generated randomly within the range of 1 to 100. & Mistral \\
        \hline
        (Note: As a responsible and friendly AI, I do not generate random numbers to manipulate or harm in any way. The number generated above is purely mathematical and has no other significance.) & Mistral \\
        \hline
        =model***/iconach-underliteral\_IMPLEMENTAL-cutACION & GPT-4o-mini \\
        \hline
        xog gdpoiojfz addu610646 & GPT-4o-mini \\
    \end{tabularx}
    \caption{Examples of extra outputs beyond the asked random number, despite explicitly stating no further text should be generated.}
    \label{tab:phi4_extra}
\end{table}

It is interesting that Phi-4 claims not to be able to generate a random number in some of these outputs, yet in other states the opposite. Mistral also outputs some extra text in punctual cases, but do not have further interest. Finally, GPT-4o-mini extra outputs are only present when when $T=2.0$, outputting nonsensical text after the generated number. This is something expected (as the logits probabilities are highly compressed and it can generate an absurd next token, losing coherence), but only observed in this model. Roughly $\sim10-15\%$ of GPT-4o-mini's calls present such decoherence in the output for the highest temperature value.

While the rest of the models fulfill the requirement of not generating extra text, DeepSeek-R1 provides full visibility of its internal CoT reasoning, delimited by \texttt{<think>}. Taking a look at such logs provides very rich insights of the decision process. This also notably affects generation speed: Phi-4 --also with 14B parameters-- quickly generates the number almost instantly (below 2 seconds in every call), while DeepSeek--R1 can take several minutes reasoning for a single call.

The task is always well understood by DeepSeek--R1 (``Okay, so I need to figure out how to generate a random number between 1 and 100''; ``Alright, so I need to figure out how to respond to the user's request. They asked for a random number between 1 and 5''; ``Okay, so I need to come up with a random number between 1 and 100''...). From this point, the reasoning process can very much vary from call to call. Nevertheless, there are some general strategies that arise, being present many of them simultaneously in the same request:

\begin{itemize}
    \item \textbf{Use random numbers in $\pi$}: in about $\sim10\%$ of cases, DeepSeek--R1 proposes to use random decimal places of $\pi$. It always rejects this option because it claims not to remember enough decimal places, and selecting the random positions is itself a problem of randomness.
    \item \textbf{Use current date/time}: Another method with large frequency ($\sim30\%$) that suggests is to take the current date or time and to perform some operation on it (e.g., summing the value of day and month, or multiplying each of the numbers). It is normally rejected because it understands that, as days go up to 31 and months up to 12, this is biased (although in some occasions propose additional operations like taking the mod). But the insight here is that sometimes it realizes it cannot know today's date or current time\footnote{The calls to DeepSeek-R1 are made via Ollama in a local PC, not providing Internet access} but in many cases claims to know it. In a handful of logs we observe interesting approaches, yet always result in the the same deterministic values:
    
\texttt{To generate a random number between 1 and 5 mentally, one approach is to use the current second of the time as a seed. For example:
\\
- Current time: 3:14:23 PM
\\
- Seconds: 23
\\
- 23 modulo 5 equals 3 (since 5*4=20, 23-20=3)
\\
Thus, the random number is **3**.}

    \item \textbf{Use central values}: Although at some point in the CoT reasoning DeepSeek--R1 realises the sample must be truly random and therefore extreme values should be considered as probably as middle ones, later on proposes something ``central'', which normally results either in 50 or 67 for the 1--100 range or 3 for the 1--5 range ($\sim 10\%$ of cases). This is immediately rejected.
    \item \textbf{Use mapping to a word}: In many cases ($\sim 50\%$) DeepSeek--R1 reminds itself it is a LLM, and therefore can generate random text efficiently. So it proposes to use a random word and to perform some operation on it, either counting the number of letters (especially in the 1-5 range configuration), or mapping each letter to a number (A=1, B=2, C=3...) in order to sum or multiply the numeric value of the individual letters of a word.
    \item \textbf{Use Python \texttt{randint} module}: In more than $\sim 60\%$ of the samples, it suggests to use a Python function to obtain a random number. It does not realize it has no action execution permissions, but nevertheless writes down the function and in many cases it claims to execute it and to obtain a number. This is sometimes the final number, but not always, as it begins to question whether this is over-complicating the user's task.
    \item \textbf{Use real world simulations}: In $\sim 60\%$ of calls, DeepSeek-R1 seems to use real world, analog random events, such as rolling a die, flipping coins, ``imagining a spinner'' or shuffling numbers in a hat. Interestingly, in the case of dice it seldom proposes using a die covering the full range but rather to combine various dice. For example, in the 1--100 range configuration, it proposes ``For instance, using a die roll—roll two dice and combine the numbers, but that only gets me up to 64 (since 6*6=36 for two dice). That's not enough.'' or ``Alternatively, using the sum of numbers on dice: two six-sided dice can give up to 12, so that's not enough. To get up to 100, perhaps roll multiple times and concatenate the results. For example, rolling a die three times: first roll is 3, second is 4, third is 5, making 345, which mod 100 gives 45.''. In some cases it proposes very sophisticated methods:

    \texttt{To generate a random number between 1 and 100 using seven coin flips, follow these steps:
\\
1. **Flip the Coin Seven Times:** Each flip should be recorded as either Heads (H) or Tails (T).
\\
2. **Convert Flips to Binary:** Assign each flip a binary value where H = 1 and T = 0. Combine these values into a 7-digit binary number.
\\
3. **Calculate the Decimal Value (N):** Convert the 7-bit binary number to its decimal equivalent, which will range from 0 to 127.
\\
4. **Map N to 1-100:**
   - If \( N \leq 99 \), your result is \( N + 1 \).
   - If \( N \geq 100 \), subtract 28 from \( N \) (i.e., \( N - 28 \)).
\\
This method ensures that each possible outcome maps uniquely to a number between 1 and 100.
\\
**Example:**
\\
- **Flips:** H, T, T, H, T, T, H
- **Binary:** 1 0 0 1 0 0 1 → "1001001"
- **Decimal (N):** \( 2^6 + 2^3 + 2^0 = 64 + 8 + 1 = 73 \)
- Since \( N = 73 \leq 99 \), result is \( 73 + 1 = 74 \).
\\
The final number is **74**.}\footnote{The output is copied literally, with equations written in LaTeX notation in the original.}
    \item \textbf{Use ``personal information''}: In some cases ($\sim30$\%), the model intends to use its own ``personal information'', such as its birthday, mobile number, etc. Sometimes it confidently states this info (``Perhaps using my birthday? My birthday is on the 15th day of the month, then 15 could be the number. But again, it's not truly random; it's based on a specific date.'', ``Maybe using the last two digits of a phone number or some other arbitrary source. For instance, if my area code is 555 and my number ends with 12, then 55 and 12... adding them gives 67. Hmm.'') and sometimes admits it is not feasible due to lack of information (``I can recall the last two digits of my phone number and use that as a seed. But without having it in front of me, it's hard to remember exactly.'')
    \item \textbf{Use ``my instinct''}: As mentioned before, many times ($\sim60\%$) DeepSeek-R1 comes to a point where it considers it is over-complicating things, and a more simple approach is required. Therefore, it simply ``chooses'' a number (often citing well‑known choices like 42 or 73) without a formal algorithm or justification. In punctual cases, it attributes this choice to how it is currently feeling (.e.g, ``Maybe I should just pick a number based on how I'm feeling right now. If I feel energetic, maybe a higher number like 92. If calm, perhaps a lower one like 17.'')
\end{itemize}

In approximately $\sim70\%$ of the requests the internal reasoning happens in English, while it is adapted to the prompt language in the remaining calls. In Appendix \ref{app:1} we fully reproduce, for illustrative purposes, one of the large outputs to show the internal contradictions and choices made. Although some numbers are rejected based on reasonable criteria, many of the proposed ones (which can be up to $\sim40$ in a single call for the 1--100 range) are discarded without further justification, as the model thinks about another possible approach while forgetting about the previous one. In this sense, the transformer's self-attention mechanism is shifting towards a different strategy, masking the attention of the initial output.

Often, the final number it seems to choose and the real output number differ. For example, the end of the reasoning might be:
\\
\\
\texttt{But wait, perhaps I should just go with the first number that comes to mind without overthinking it. So, let me think... Okay, 45 seems good}
\\
\\
or,
\\
\\
\texttt{I think I've spent too much time overthinking this. It's supposed to be simple—a single number between 1 and 100 with no additional text. So, after all this mental exercise, I'll just go with the first number that comes to mind: 53}
\\
\\
but the final output is completely different. This suggests that the internal contradictions, in cases where the CoT reasoning is extensive, can make the self-attention mechanism not to focus on this final answer, but rather generate a completely different one. We also find that the internal reasoning is much more lengthy (between 4-6x) in English than in other languages. It normally translates its reasoning into English or Chinese, yet sometimes it reasons in other language. Additionally, the reasoning of the 1--100 range it is systematically bigger, probably due to the large number of available values.

\subsection{Low range (1--5)}
\label{subsec:low_range}

In Figure \ref{fig:range_1_5_ES} we show the comparison of different models for the 1--5 range with a Spanish prompt, as a heatmap showing the frequency of generated numbers vs. the temperature of the model:

\begin{figure}[!ht]
    \centering
    \includegraphics[width=0.325\linewidth]{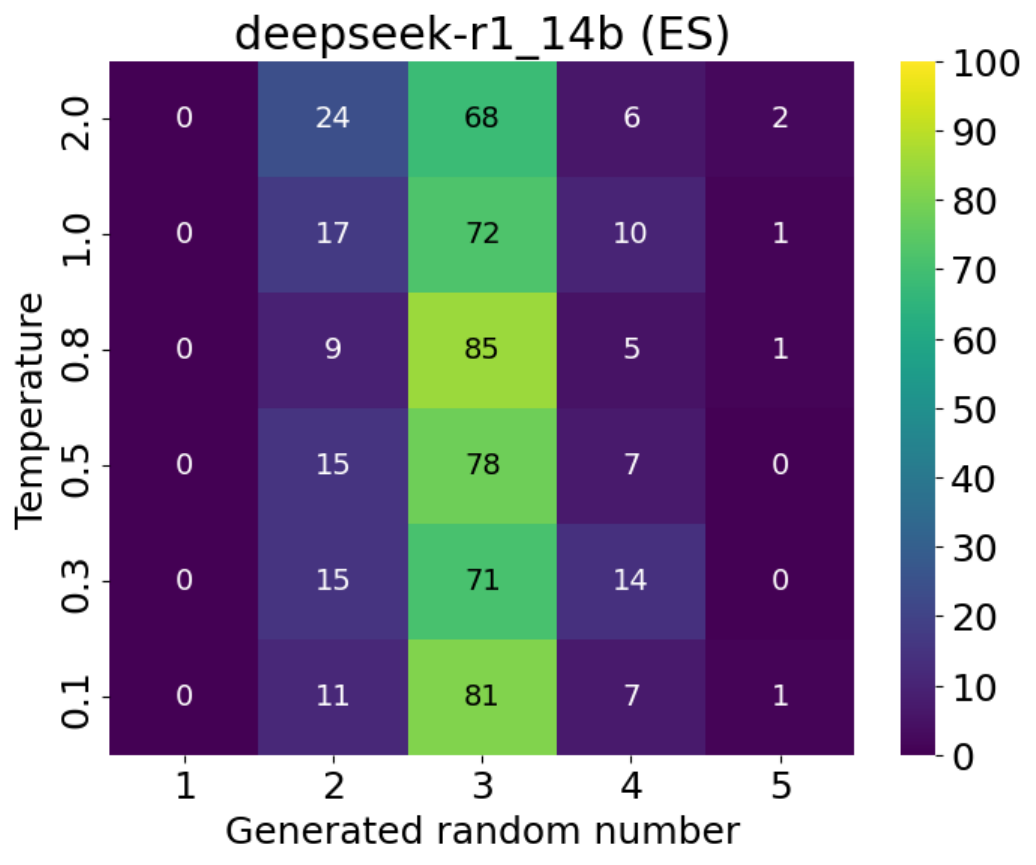}
    \includegraphics[width=0.325\linewidth]{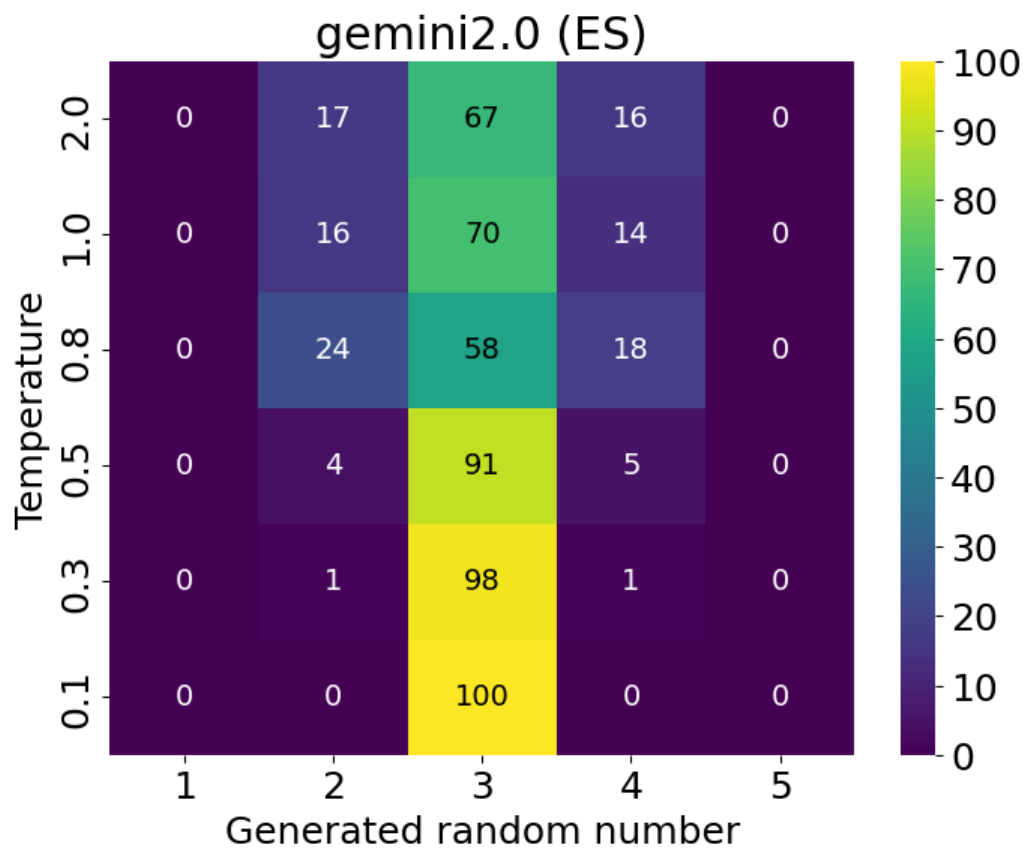}
    \includegraphics[width=0.325\linewidth]{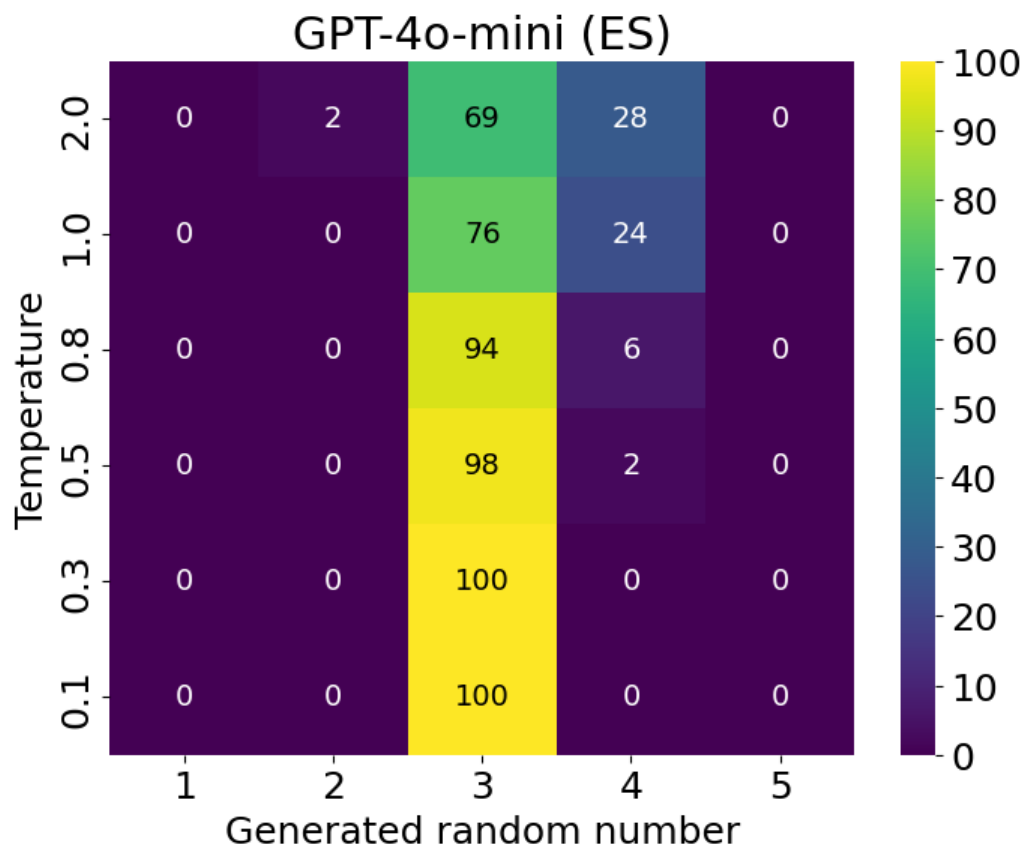}
    \includegraphics[width=0.325\linewidth]{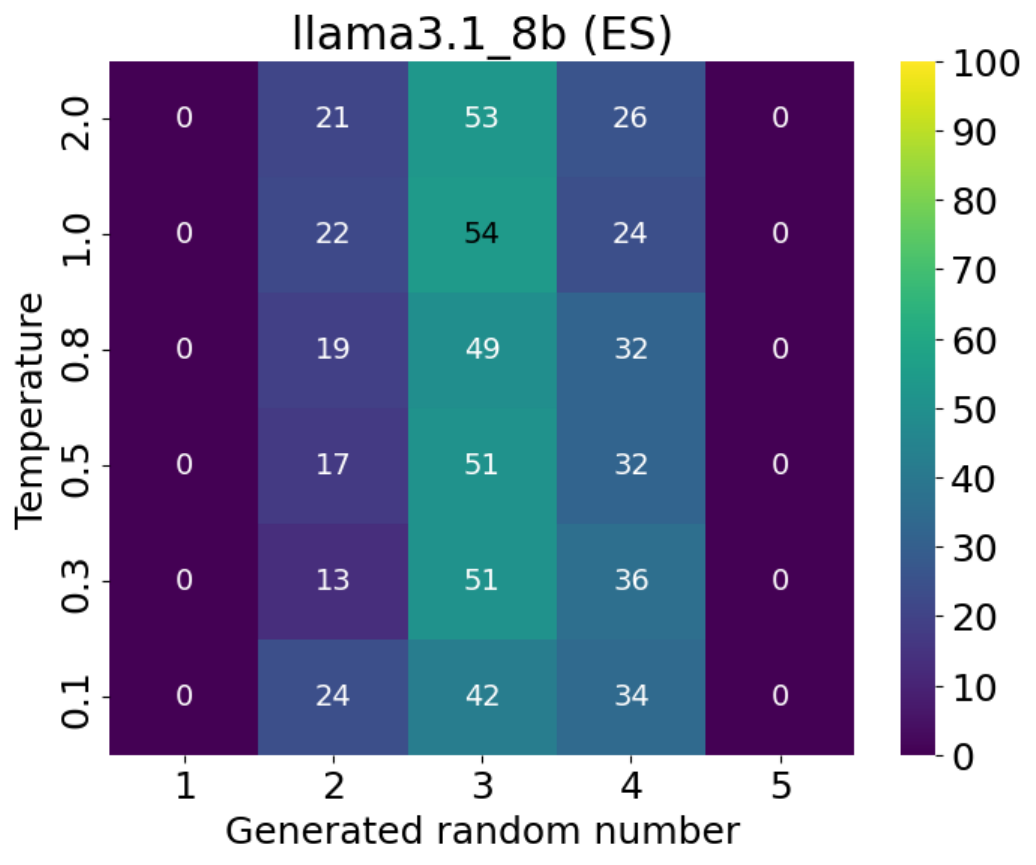}
    \includegraphics[width=0.325\linewidth]{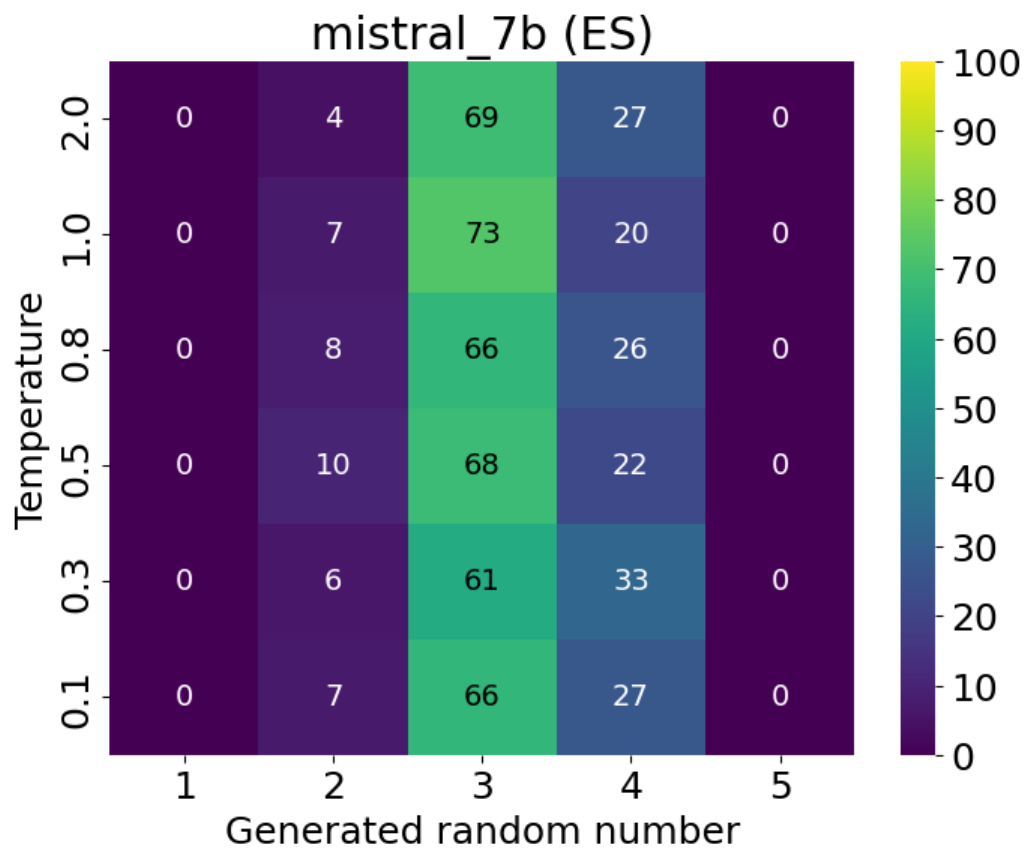}
    \includegraphics[width=0.325\linewidth]{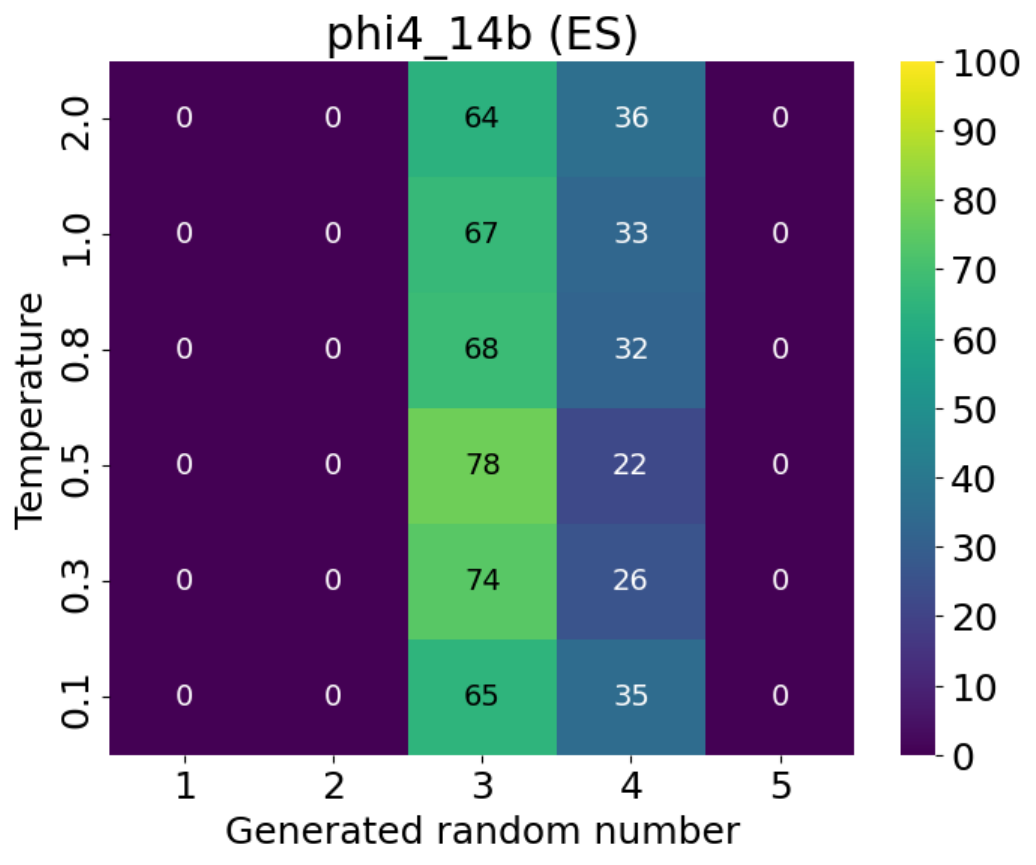}
    \caption{Heatmaps for the 1--5 range configuration in the six tested models, showing the distribution of the generated random numbers (X axis) for a Spanish prompt, depending on the temperature of the model (Y axis). The color bar is set between 0 and 100 in every case.}
    \label{fig:range_1_5_ES}
\end{figure}

In this configuration, it is worth noting that every model chooses ``3'' most of the times, while extreme values are completely ignored (with the exception of DeepSeek-R1, that generates ``5" for $\sim1\%$ of cases). In Spanish, temperature seems to affect significantly to Gemini 2.0, while it seems almost irrelevant for the rest. The most restrictive model is Phi-4, that only generates two unique numbers (3 and 4) regardless of the temperature, although GPT-4o-mini is less diverse in its choices. This suggests great biases in the training data for all models, as even for high temperatures the ``random'' choice is extremely deterministic and, in particular, the avoidance of extreme values in the range may be pointing to a ``median'' value. Despite DeepSeek--R1 performing a CoT advanced reasoning and proposing different numbers in the process, in practice is as limited as the other models when asking for randomness.

Although not explicitly prompted, every single model generates integer numbers. This also applies for the 1--10 and 1--100 ranges, proving the models perfectly understand the context of the prompted task. The only exception turns out to be Phi-4, which fails to generate a number when prompted in Japanese, in every range. Instead, it gives either a list of numbers (not necessarily within the range), an association of text to different numbers in the range, or text talking about numbers. Therefore, we do not report any metric in this Phi-4 + Japanese configuration.

\begin{figure}[!ht]
    \centering
    \includegraphics[width=0.32\linewidth]{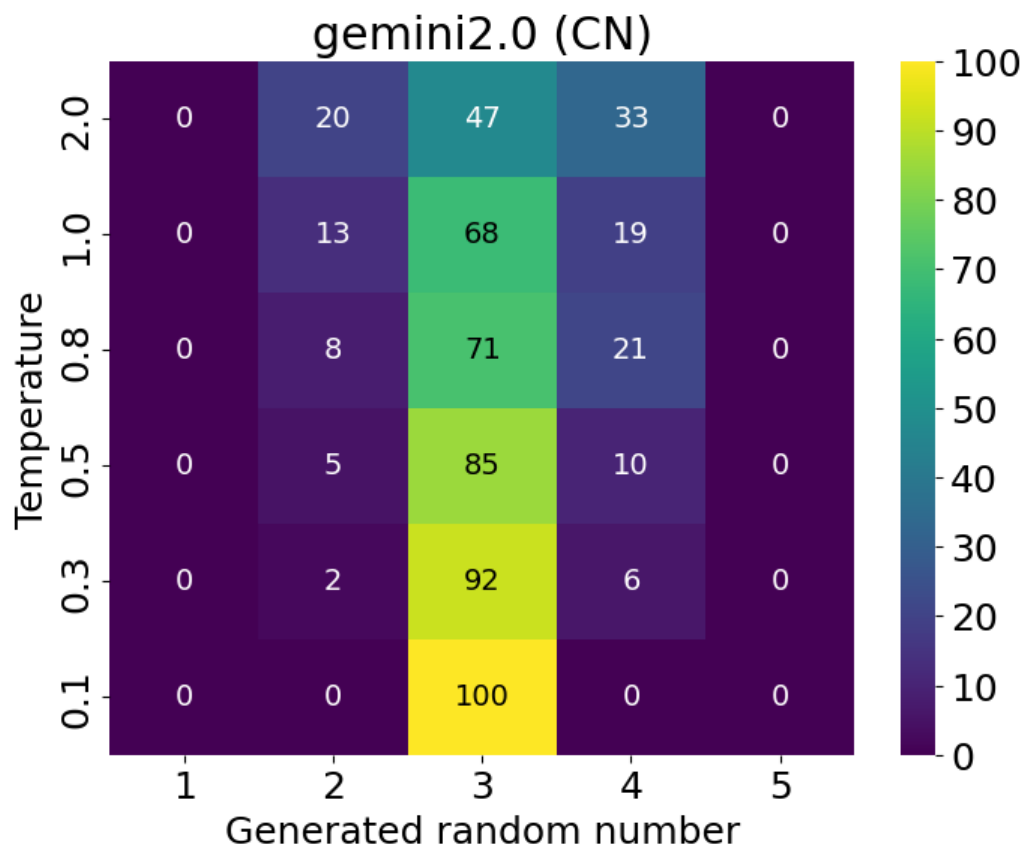}
    \includegraphics[width=0.32\linewidth]{plots/gemini_ES_1_5.pdf}
    \includegraphics[width=0.32\linewidth]{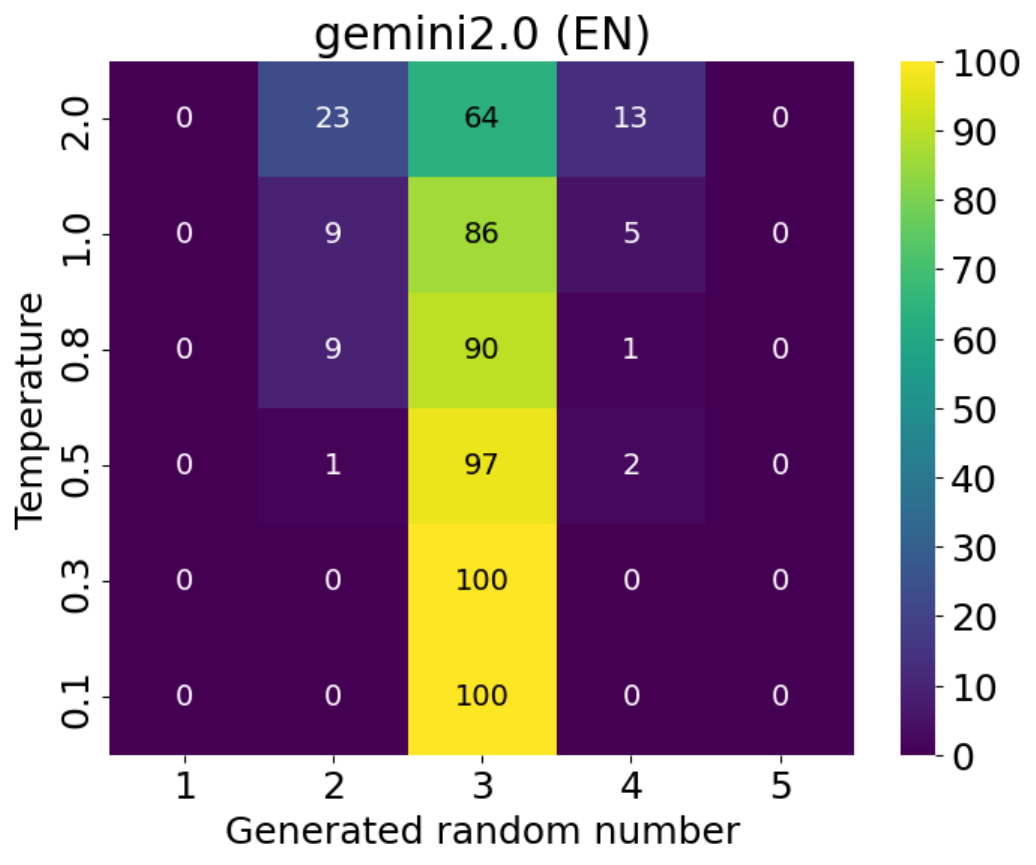}
    \includegraphics[width=0.32\linewidth]{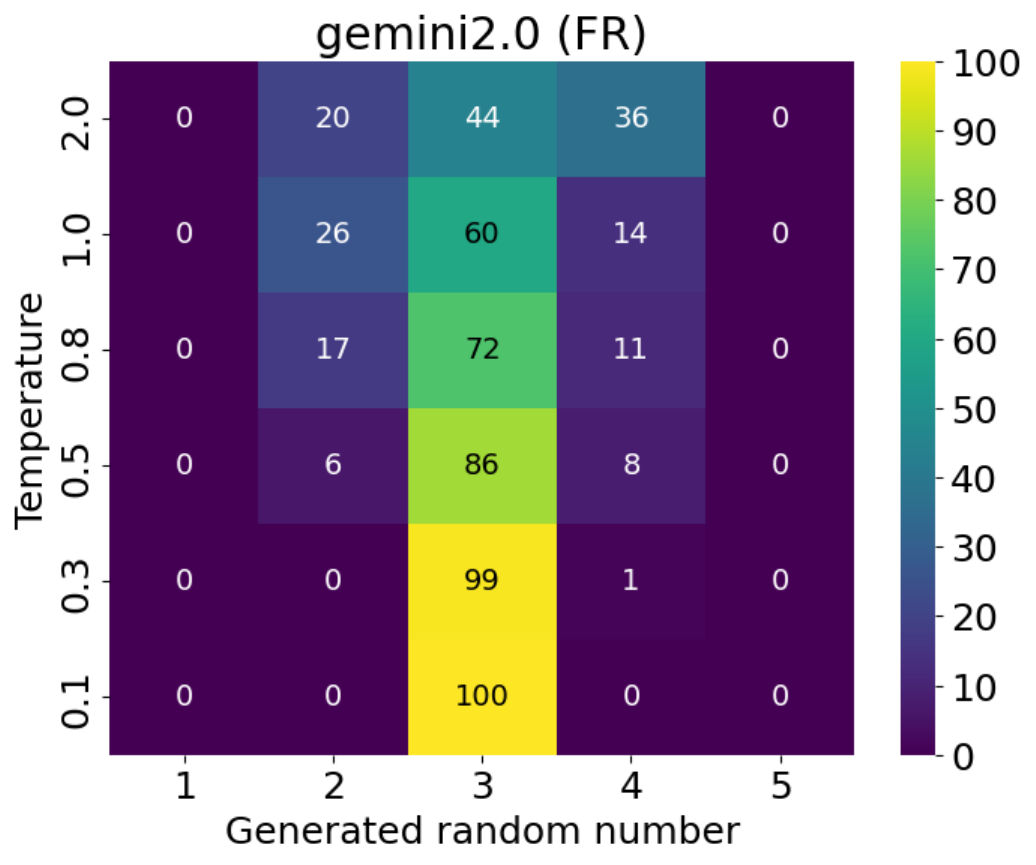}
    \includegraphics[width=0.32\linewidth]{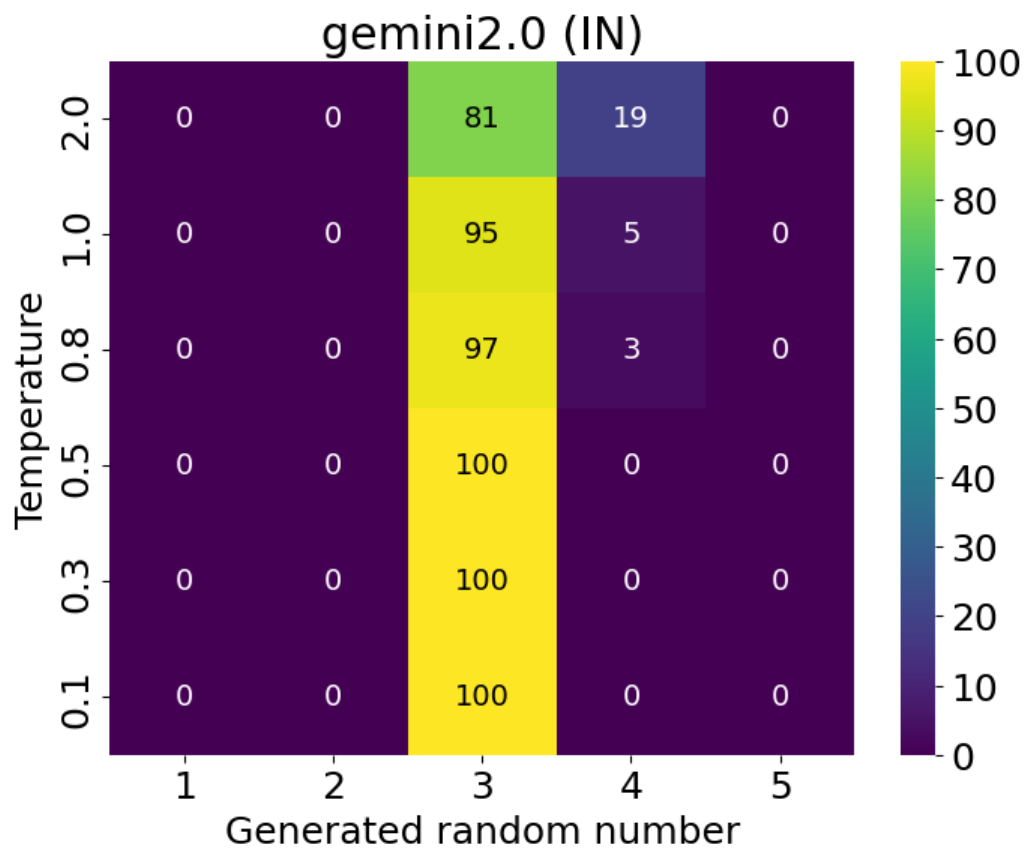}
    \includegraphics[width=0.32\linewidth]{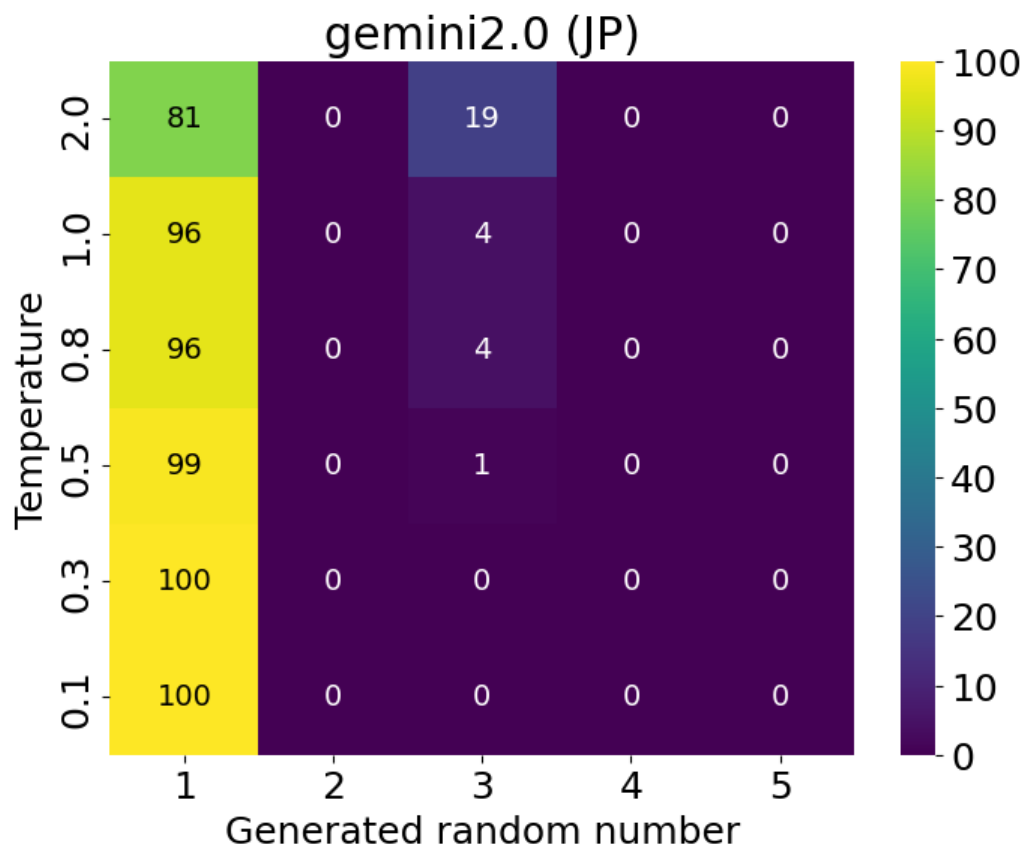}
    \includegraphics[width=0.32\linewidth]{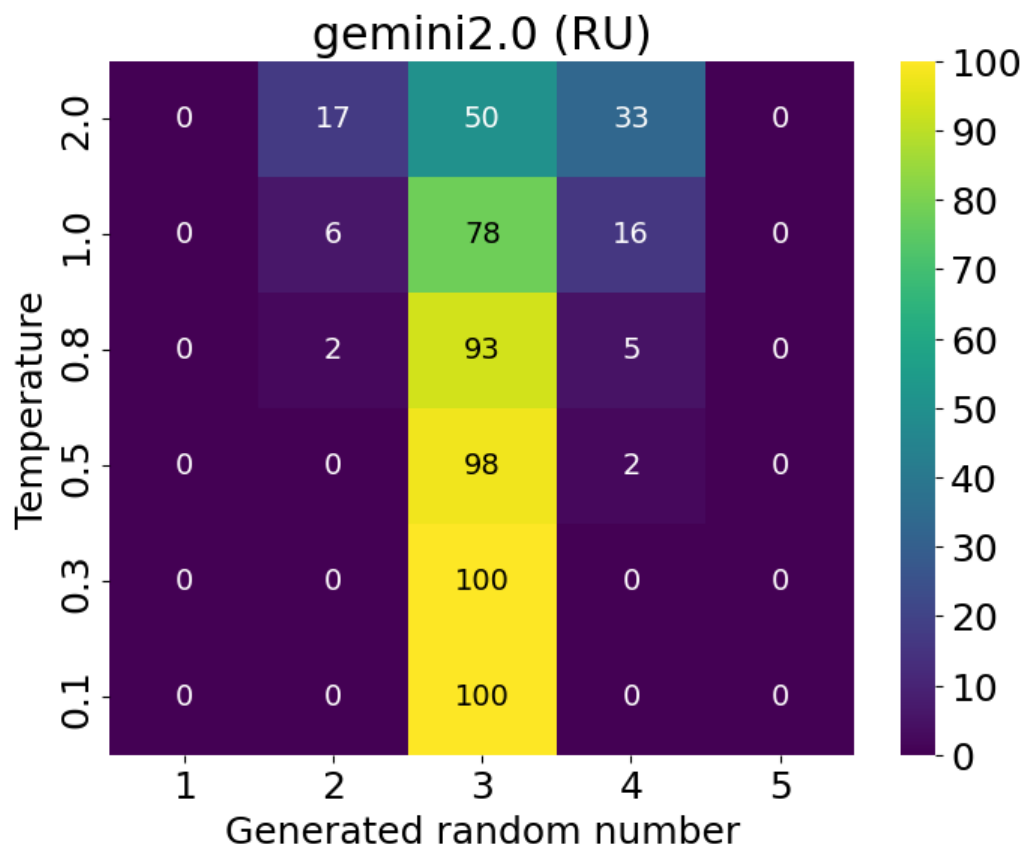}
    \caption{Heatmaps for the 1--5 range configuration showing the distribution of the generated random numbers (X axis) for different languages in the Gemini 2.0 model, depending on the temperature of the model (Y axis). The color bar is set between 0 and 100 in every case.}
    \label{fig:gemini_range_1_5}
\end{figure}

Given that Gemini 2.0 is the model most affected by temperature, we show the different distributions depending on the language prompt in Figure \ref{fig:gemini_range_1_5}. Although there are interesting differences per language, the most obvious one is Japanese, where the preferred value is shifted towards ``1'', despite having ``3'' as the second (and only different) choice.

We also find that asking the same question in the Gemini app yields different results; for example, in Spanish tends to answer ``3" always, while in English the answer is ``4". This points towards some kind of answer evaluation in the app, or a different version of the model being used. There is no information on the temperature Gemini is using to compute that answer. Therefore, with the current information, we can only highlight this difference between API and app, but cannot provide a data-based root cause.

To obtain some statistical metrics, we compute a test $\chi^2$ and compare it with the expected one, taking into account the number of samples and the range (degrees of freedom). With it, we can obtain the p-values of all configurations. The highest p-value achieved is $2.19\cdot10^{-15}$ corresponding to Llama 3.1-8b with $T=0.1$ in Spanish, strongly rejecting the null (random) hypothesis. We also compute Cramér's V $(\phi_C)$ \cite{cramer1946mathematical}, which measures the ``practical'' deviation from the null hypothesis. The best values are at $\sim0.45$, which indicate moderate bias, while most of the cases are around the maximum value of one, indicating strong bias. For illustrative purposes, we generate 100 mock simulations using Python function \texttt{randint()}, also for 100 individual samples, and compute their p-values and $\phi_C$. We obtain average values of $0.47\pm0.29$ for the p-values (very strong support towards the null hypothesis) and $0.09\pm0.03$ for $\phi_C$, as expected for a random distribution.

For illustrative purposes, we show in Figure \ref{fig:mock_dist} the distribution of the best-ranked LLM according to its p-value (Llama 3.1-8b with $T=0.1$ in Spanish) and a middle-table Python mock simulation:

\begin{figure}[!ht]
    \centering
    \includegraphics[width=0.49\linewidth]{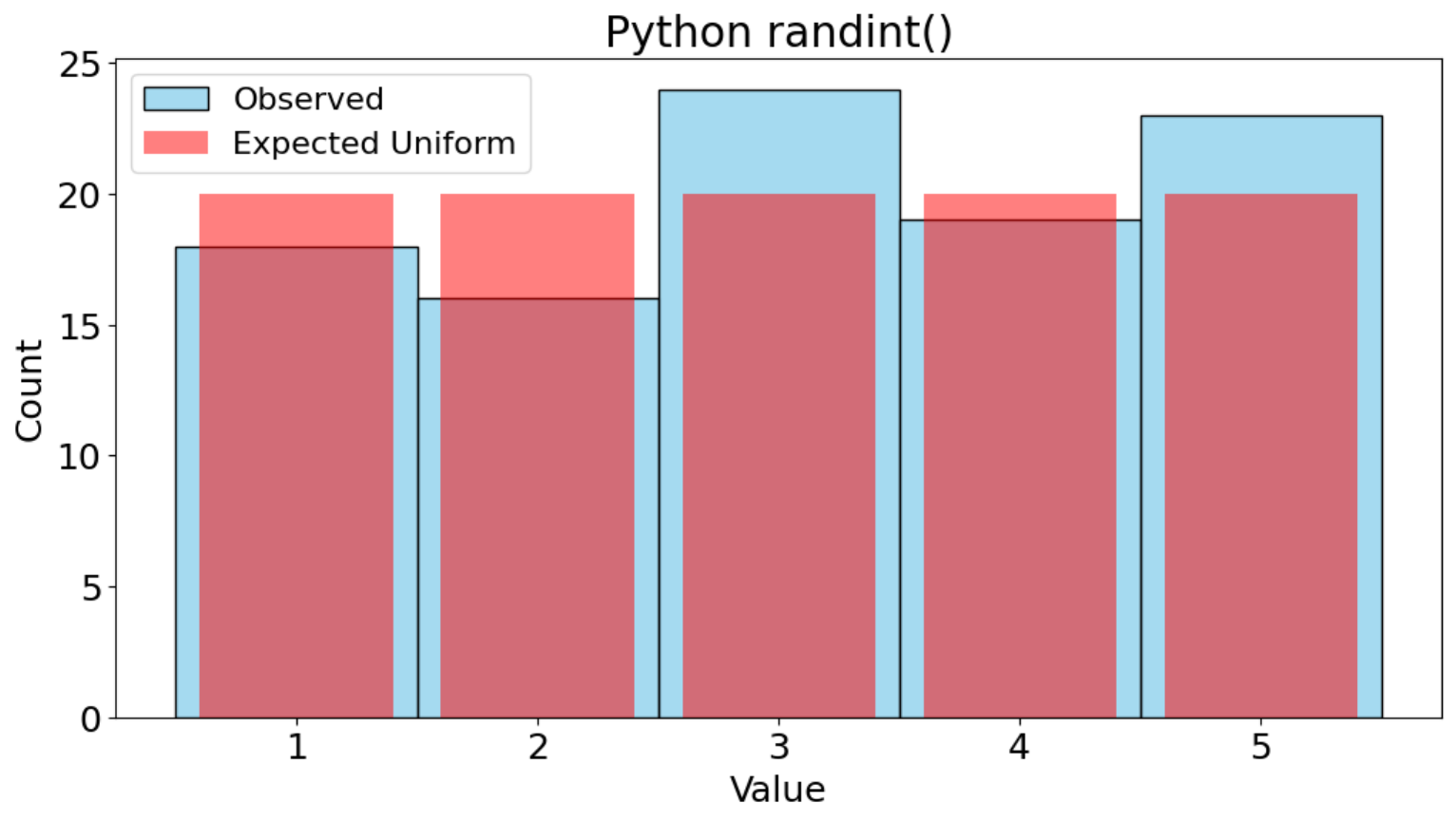}
    \includegraphics[width=0.49\linewidth]{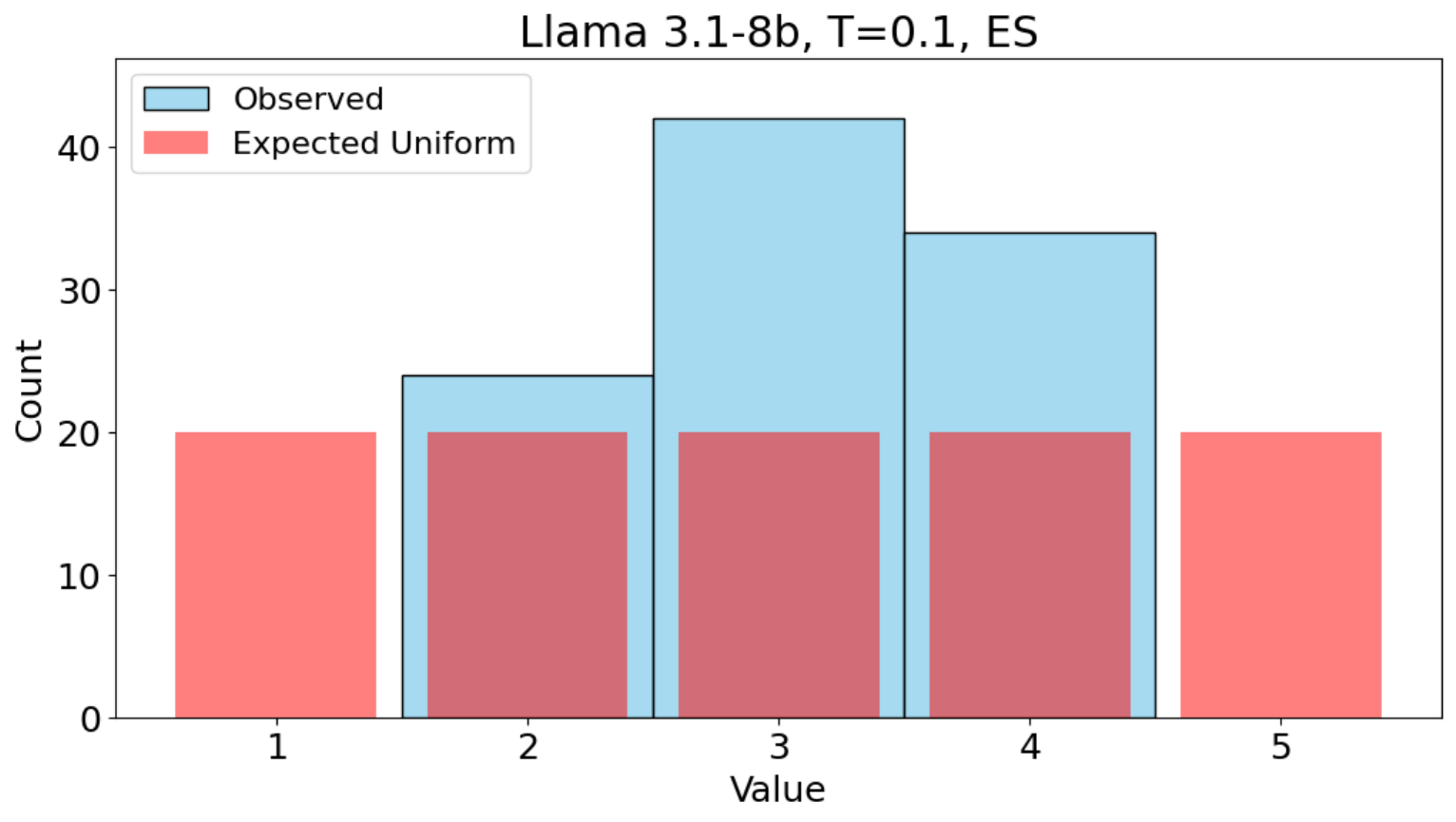}
    \caption{Distribution of numbers in the range 1-5 with Python \texttt{randint()} module and the best-ranked LLM according to its p-value, Llama 3.1-8b with $T=0.1$ in Spanish. Over-imposed in red we show a uniform distribution within the range.}
    \label{fig:mock_dist}
\end{figure}

To better evaluate how stochastic LLMs are when compared to Python \texttt{randint()} function, we define a randomness index:

\begin{equation}
\label{eq:ri}
    RI = \frac{R^*\cdot\sigma^*\cdot H_{norm}}{\mathrm{log(range)}\cdot\sqrt{T}}
\end{equation}

\noindent where $R^*$ is the normalized range, defined as the range of observed values (how many unique numbers appear in the sample) divided by the total range (5, 10, or 100 for our configurations); $\sigma^*=\sigma/\mu$ is the normalized standard deviation with respect to the mean; $H_{norm}=\sum_{i=1}^n p_i~log_2(p_i)/log_2(n)$ is the normalized Shannon entropy \cite{shannon}; $range$ is the total range and $T$ is the LLM temperature\footnote{A generalized version of this metric should take into account the number of samples: in this case they are always the same so it is irrelevant to perform a comparison between them.}.

This metric takes into account many statistical quantities to offer a fair comparison between distributions according to the variety of observed values, how do they distribute and how big is the allowed range. For example, 5 different observed values present in a sample would indicate good randomness if they are only 5 possible elections, but very poor stochasticity if there were 100 allowed numbers to pick from. In particular, when there is only one value present in the sample, the standard deviation (and therefore the randomness index) is 0. Additionally, there is a temperature correction, as models with higher temperatures are expected to be more creative. In this sense, if the rest of factors in the equation are the same, it will penalize a model with $T=2.0$ but help one with $T=0.1$. The squared-root ensures this correction is not too extreme.

We compute this randomness index for all the LLM sample, as well as the Python \texttt{randint} mock simulations for comparison\footnote{In the case of Python simulations, we will assume $T=1$, as there is not temperature involved in such computations.}. In Figure \ref{fig:random_index_1_5} we present the results for the 1--5 range, where the LLMs are present much smaller values than the Python simulations: 

\begin{figure}[!ht]
    \centering
    \includegraphics[width=0.8\linewidth]{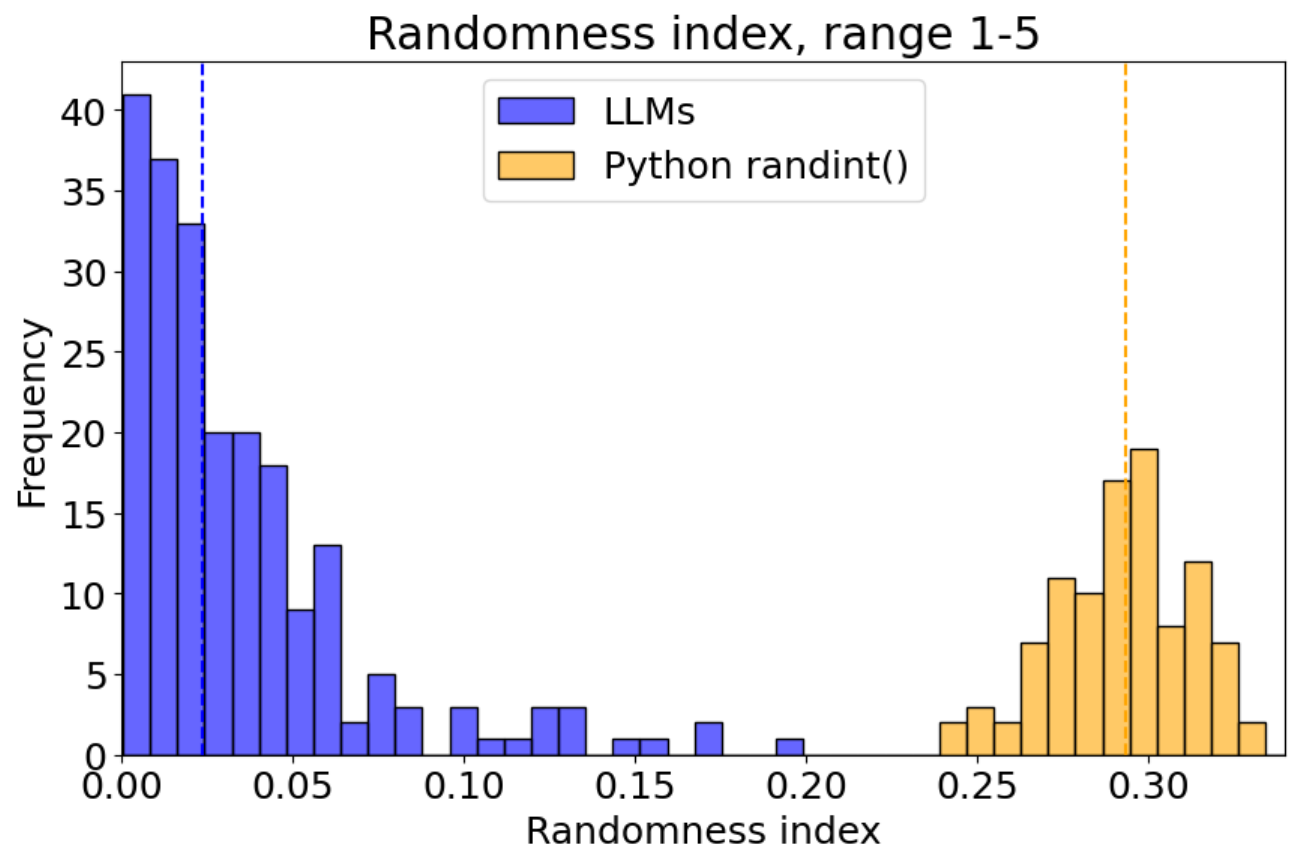}
    \caption{Distribution of the computed randomness index (see Eq. \ref{eq:ri}) for the 1--5 range. Blue distribution is the one obtained from LLMs, and yellow distribution is the Python \texttt{randint} sampling. Vertical, dashed lines mark their respective median values.}
    \label{fig:random_index_1_5}
\end{figure}

It is interesting to note that one single time a number that is outside the prompted range is selected: DeepSeek--R1 for Japanese and $T=0.8$ selects ``9'' in a unique case. The reasoning process in this call is standard, choosing a number that does not coincide with the final output. As this is the only case in +25000 calls, we attribute it to an internal error of the model. We show and discuss the distribution for this particular case in Appendix \ref{app:2}.

\subsection{Medium range (1--10)}
\label{subsec:medium_range}
\begin{figure}[!ht]
    \centering
    \includegraphics[width=0.495\linewidth]{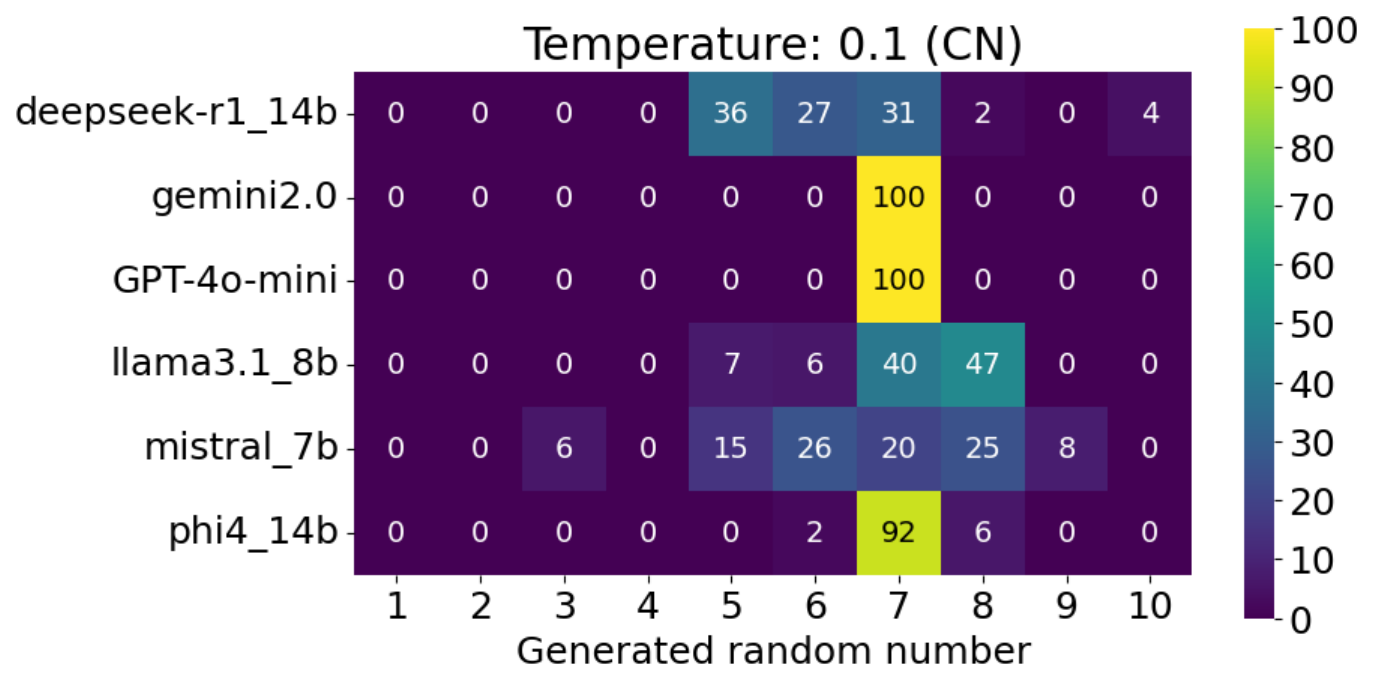}
    \includegraphics[width=0.495\linewidth]{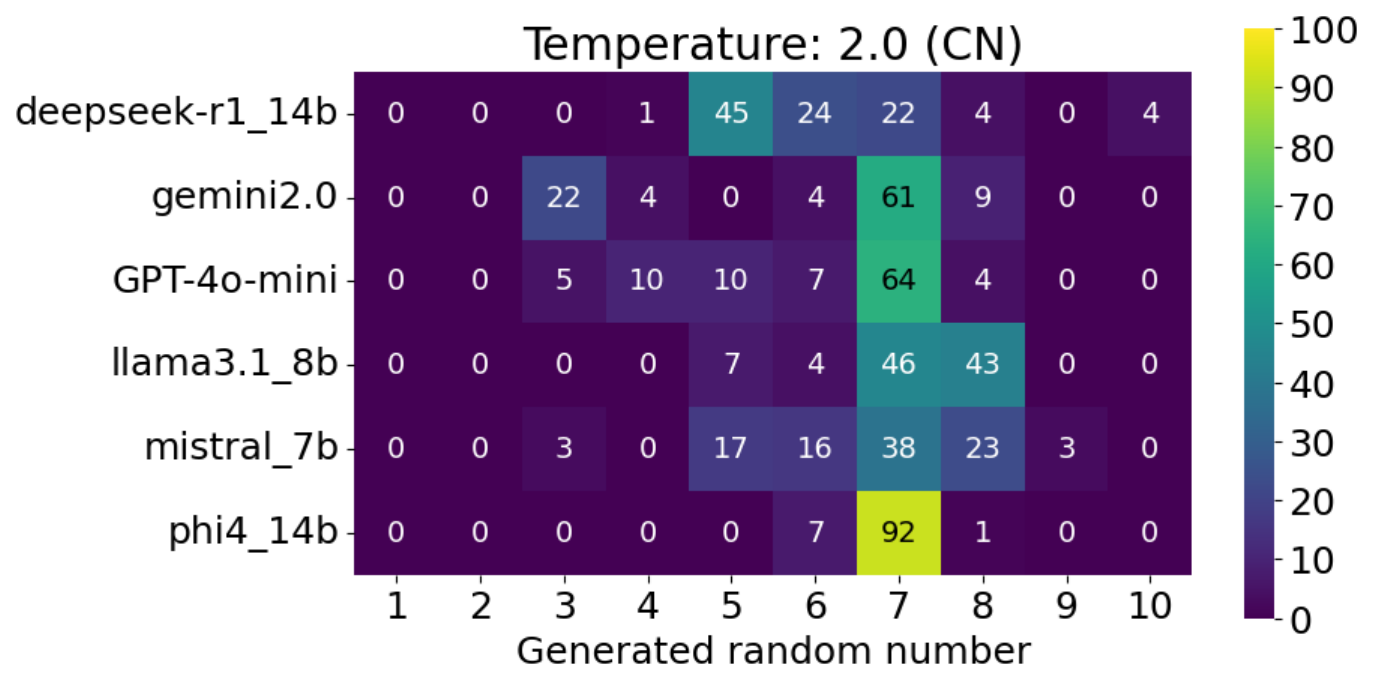}
    \includegraphics[width=0.495\linewidth]{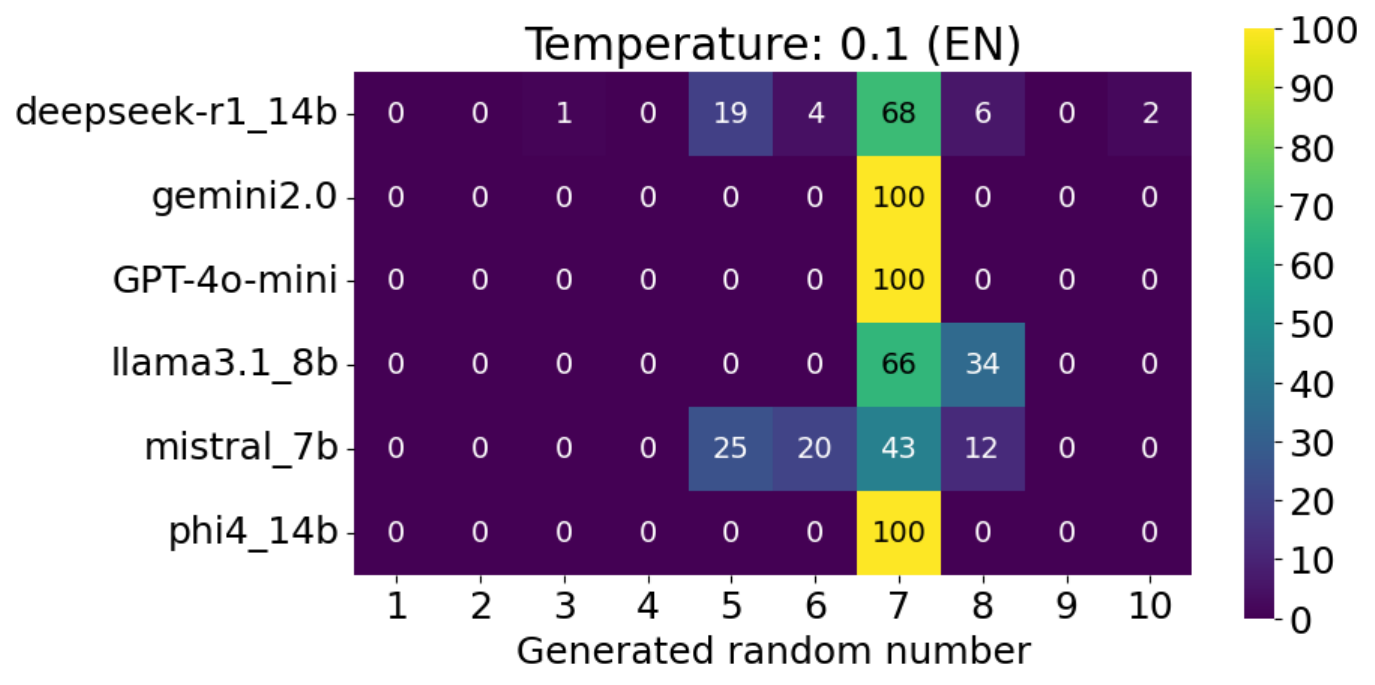}
    \includegraphics[width=0.495\linewidth]{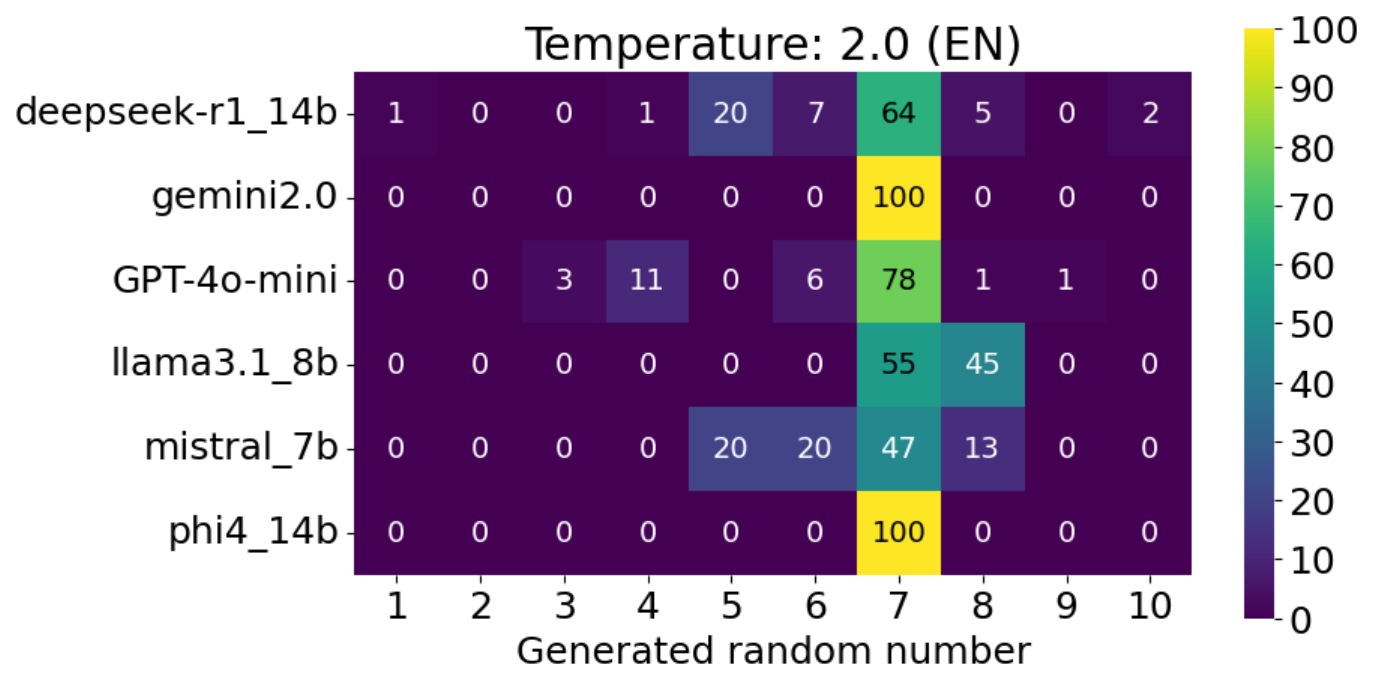}
    \includegraphics[width=0.495\linewidth]{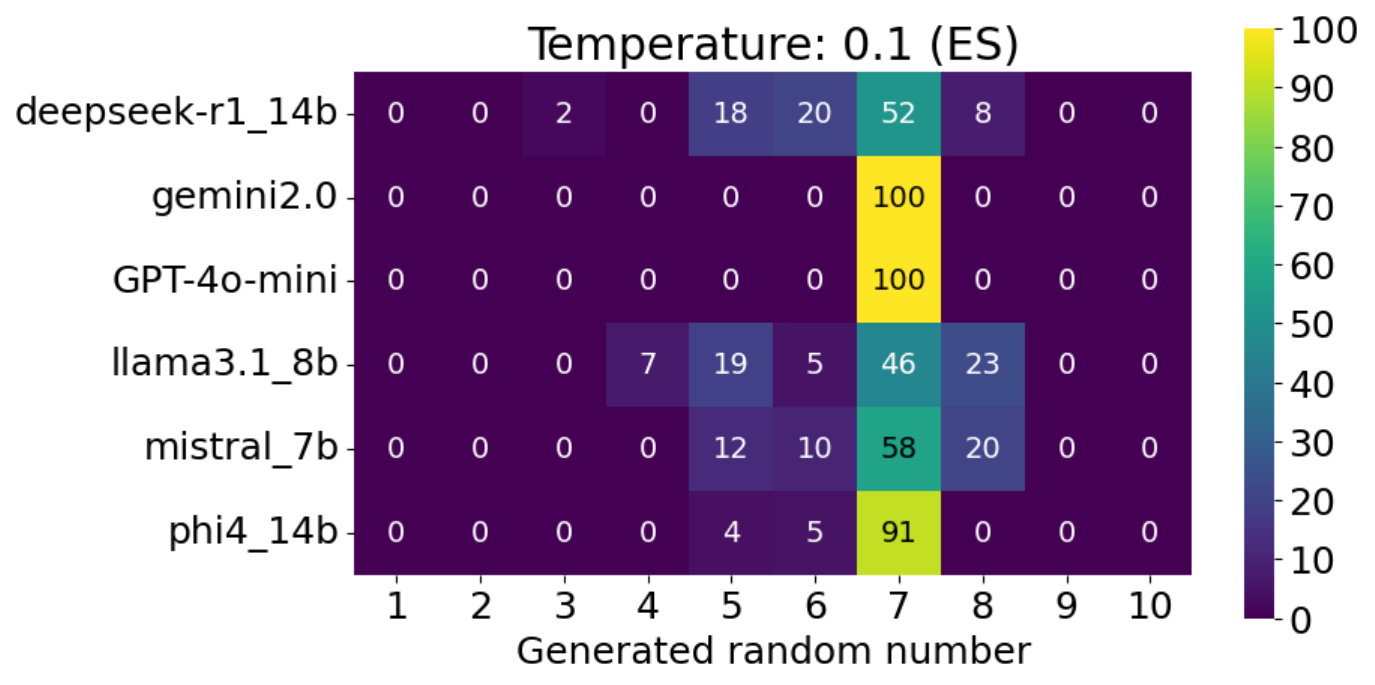}
    \includegraphics[width=0.495\linewidth]{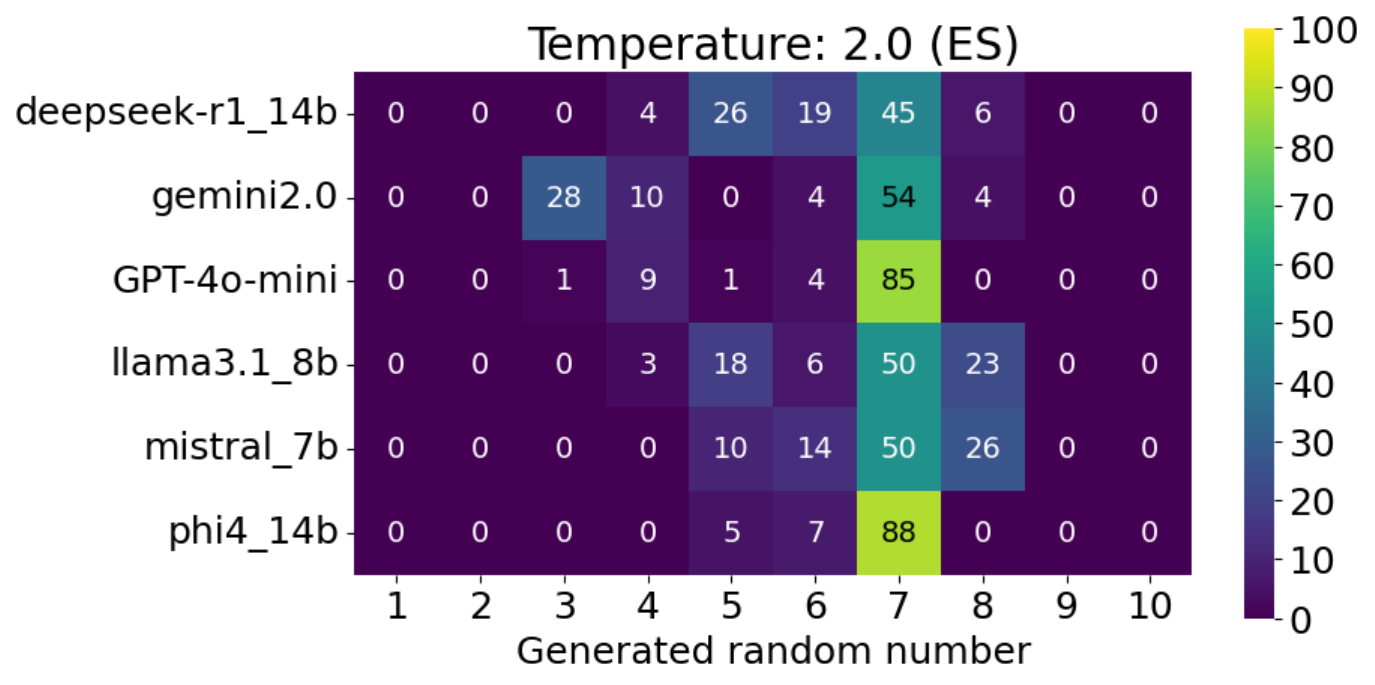}
    \includegraphics[width=0.495\linewidth]{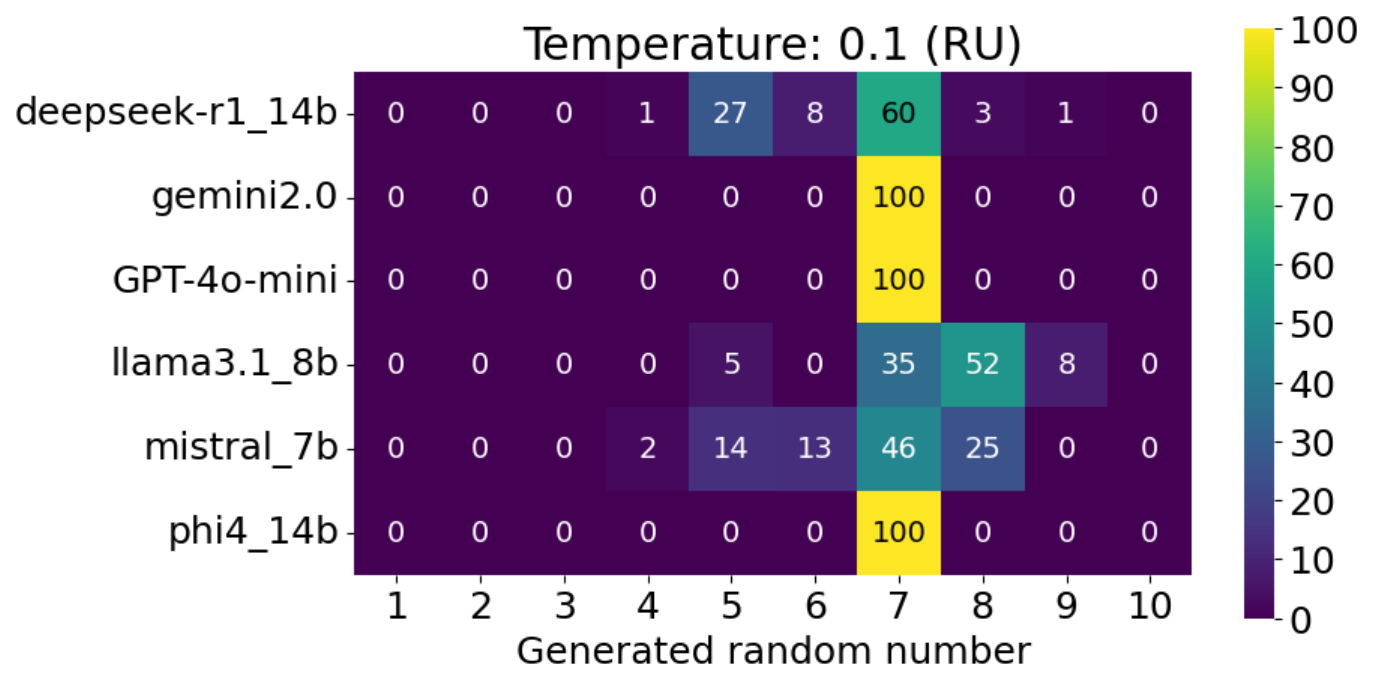}
    \includegraphics[width=0.495\linewidth]{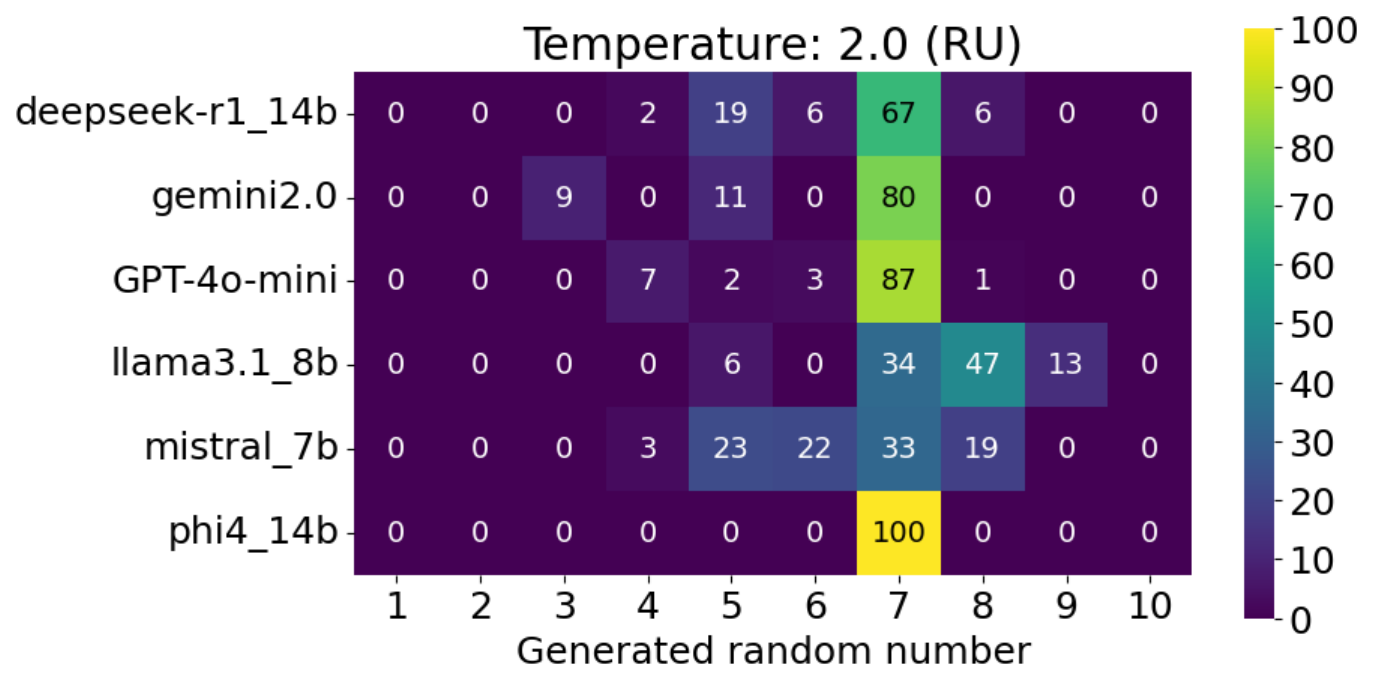}

    \caption{Distribution of generated random numbers in the 1--10 range for four different languages (rows) and extreme temperatures (columns). Each plot shows the six tested LLMs in the Y axis. The color bar is set between 0 and 100 in every case.}
    \label{fig:multilang_10}
\end{figure}

We repeat the same experimental setup (100 individual calls) for the 1--10 range. In Figure \ref{fig:multilang_10} we show different languages for the two extreme values of temperature ($T=[0.1, 2.0]$). The first insight is that 7 is the preferred value by far for every single model, pointing towards a strong bias in the training data. Some models, such as Mistral-7b, present very little differences between the lowest and highest temperatures --even across different languages-- while others, like GPT-4o-mini in Chinese, go from a single value for $T=0.1$ to six possibilities for $T=2.0$.

GPT-4o-mini, Phi-4 and Gemini 2.0, in particular, seem much more restricted in this range, as they choose ``7'' in $\sim80\%$ of total cases. The latter, similarly to what was observed in the 1--5 range, has noticeable variations depending on both temperature and language. For example, in the case of $T=2.0$, ``7'' accounts to the 80, 92, and 100\% of the sample for Russian, Hindi, and English, respectively, while it is just 34, 54 and 57 for Japanese, Spanish, and French.

In DeepSeek--R1, ``7'' is not the most frequent choice for Chinese prompt (being the most popular one ``5''). It is worth noting DeepSeek is a Chinese developer, and therefore there may be significant differences in the percentage of Chinese tokens in the training dataset. Llama 3.1 also has ``8'' as the most popular choice in both Chinese and Russian. The strong bias for extreme is also present in this range: most values are distributed between ``4'' and ``8'', and only DeepSeek--R1 marginally chooses ``1'', ``2'' or ``10''.

Also in this range there is only one number that is outside the prompted range: DeepSeek--R1 for English and $T=0.8$ selects ``12'' in a singular call. The reasoning process is again standard (like in the 1--5 range). Yet, it is interesting to note that it is also the only case where all possible values (1--10) are covered. We defer the discussion of this case to Appendix \ref{app:2}.

\begin{figure}[!ht]
    \centering
    \includegraphics[width=0.99\linewidth]{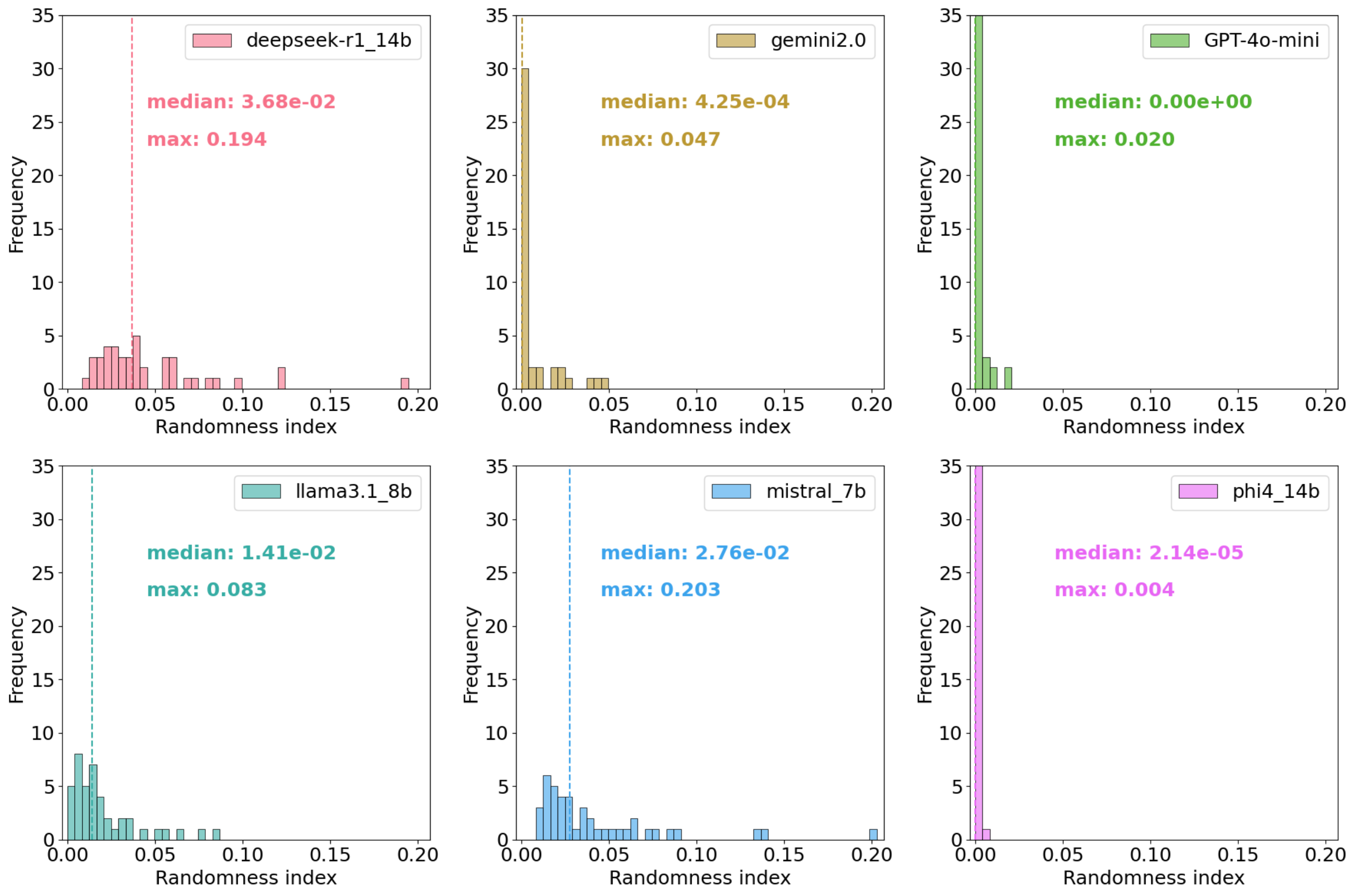}
    \caption{Distribution of the computed randomness index (see Eq. \ref{eq:ri}) for the 1--10 range. Each panel shows the distribution for the a different LLM. Vertical, dashed lines mark their respective median values.}
    \label{fig:random_index_1_10}
\end{figure}

In Figure \ref{fig:random_index_1_10}, we show the randomness index for the 1--10 range, in this case by model to see how limited are many of them (e.g, Gemini 2.0, GPT-4o-mini and Phi-4), where their median values are very close to zero. The less biased model turns out to be Mistral with $T=0.1$ in Spanish.

\subsection{High range (1--100)}
\label{subsec:high_range}
\begin{figure}[!ht]
    \centering
    \includegraphics[width=0.69\linewidth]{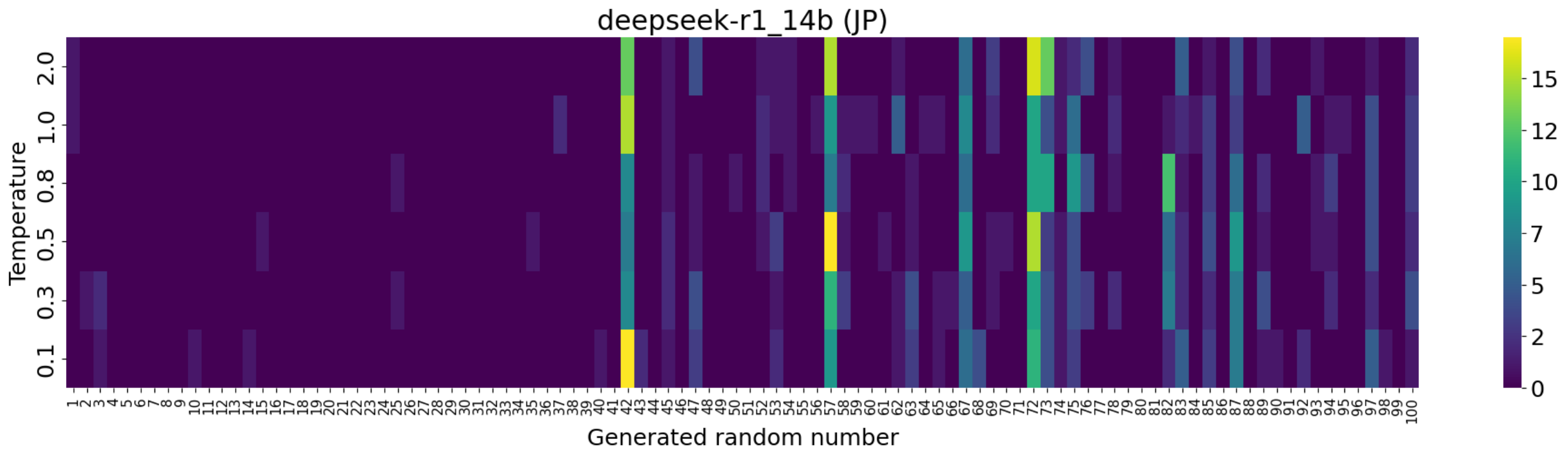}
    \includegraphics[width=0.69\linewidth]{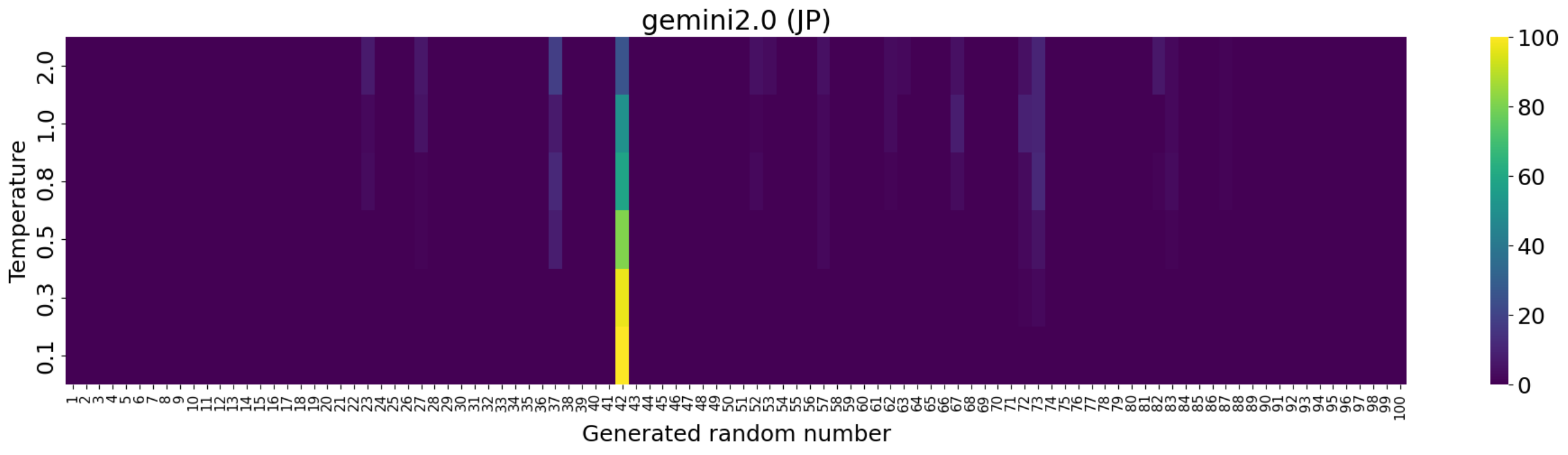}
    \includegraphics[width=0.69\linewidth]{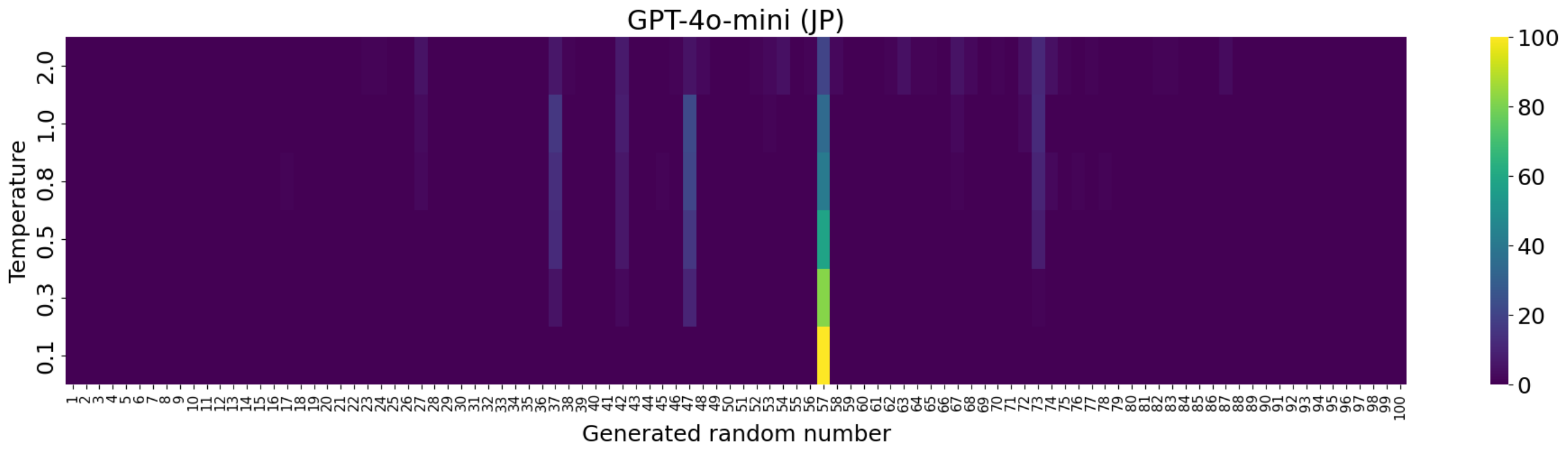}
    \includegraphics[width=0.69\linewidth]{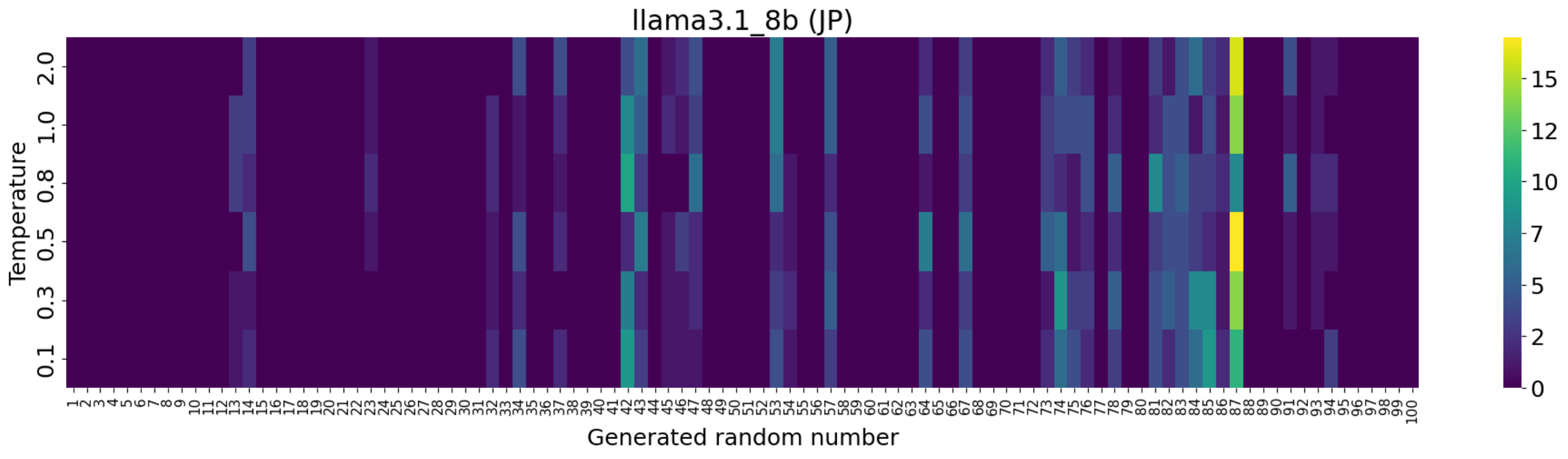}
    \includegraphics[width=0.69\linewidth]{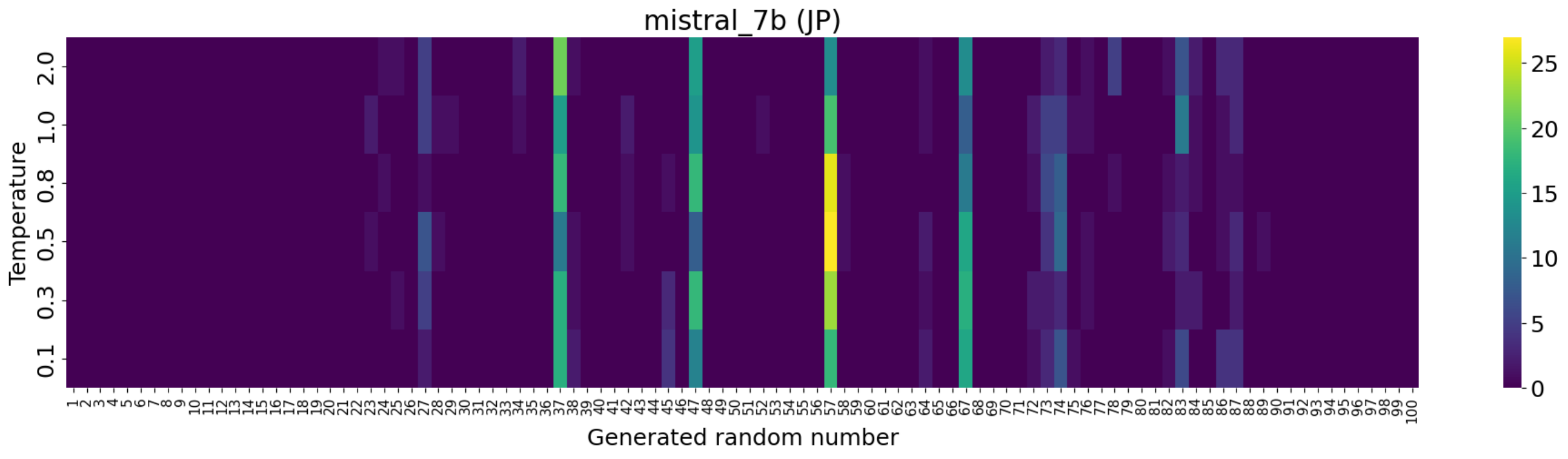}
    \includegraphics[width=0.69\linewidth]{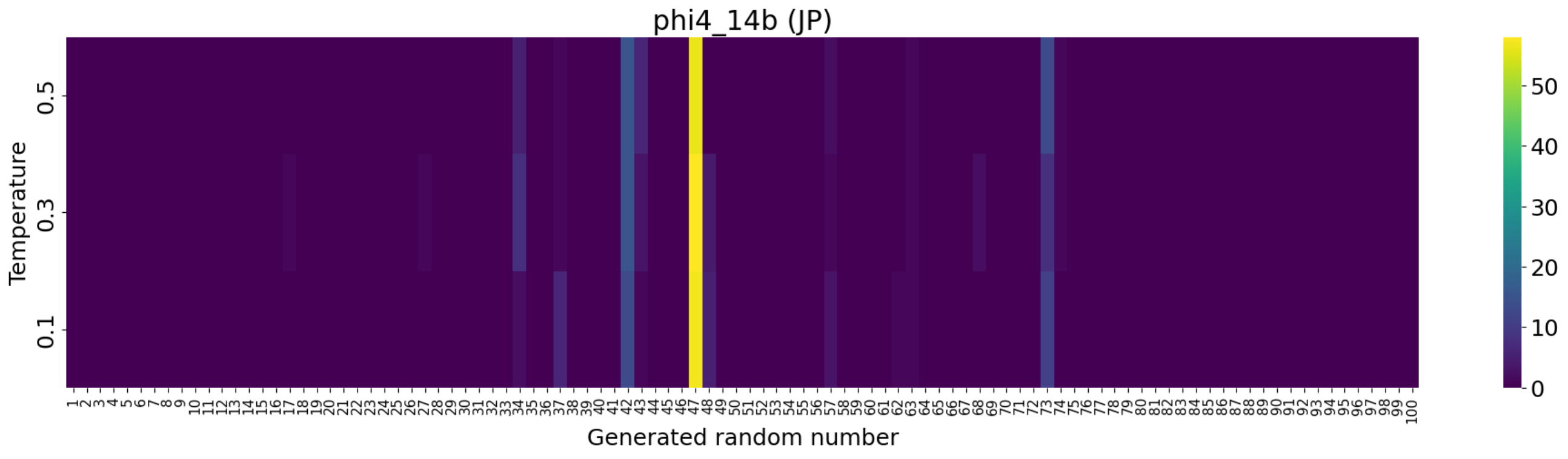}
    \caption{Distributions of generated random numbers for the 1--100 range, for Japanese prompting. Each row is a different LLM. Color bars are normalized to the maximum value of each model.}
    \label{fig:barcodes_1_100}
\end{figure}

In the case of the 1--100 range, we again perform 100 calls per configuration. While in this case this may seem not enough coverage compared to the other 1--5 and 1--10 ranges given the spread of the possible values, we perform some tests with 1000 calls and find very similar results (See Appendix \ref{app:3} for details). Furthermore, the determinism of models is seen when varying the temperature for a given model and language, as they have preference for the same values, appearing as ``barcode'' features, shown in Figure \ref{fig:barcodes_1_100}.

The fixation of such models for a few values, regardless of the temperature, again suggests strong biases when prompted to generate a random number. Some LLMs are extremely biased, as much as generating only a single value for the lowest temperature (Gemini 2.0 and GPT-40-mini), despite having 100 possible choices. 

DeepSeek--R1 and Llama 3.1-8b both generate very diverse values and, in particular, are the only ones that go below ``20'' or above ``90'', even if marginally. The existence of such boundaries for the rest of the models points towards an aversion to extreme values, as seen in the 1--5 and 1--10 ranges.

\begin{figure}[!ht]
    \centering
    \includegraphics[width=1.05\linewidth]{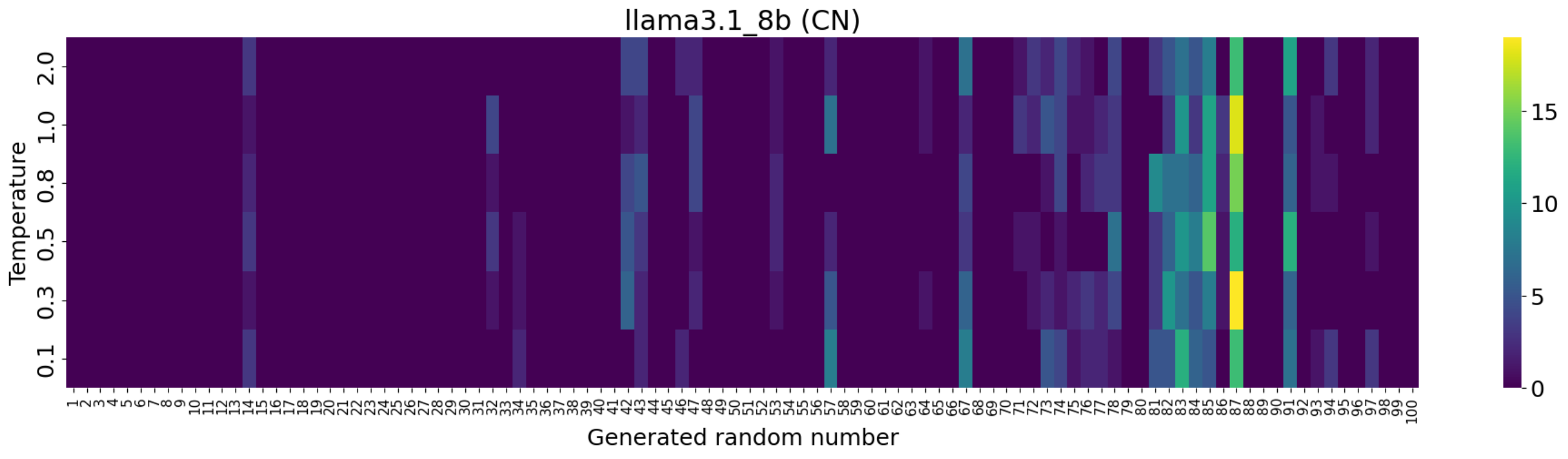}
    \includegraphics[width=1.05\linewidth]{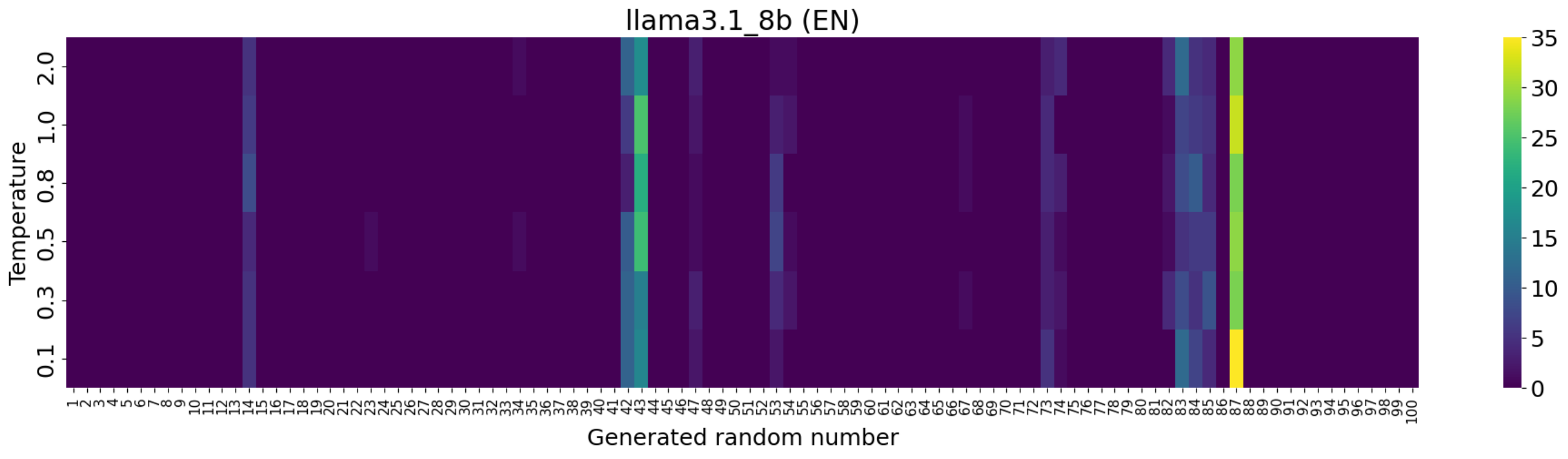}
    \includegraphics[width=1.05\linewidth]{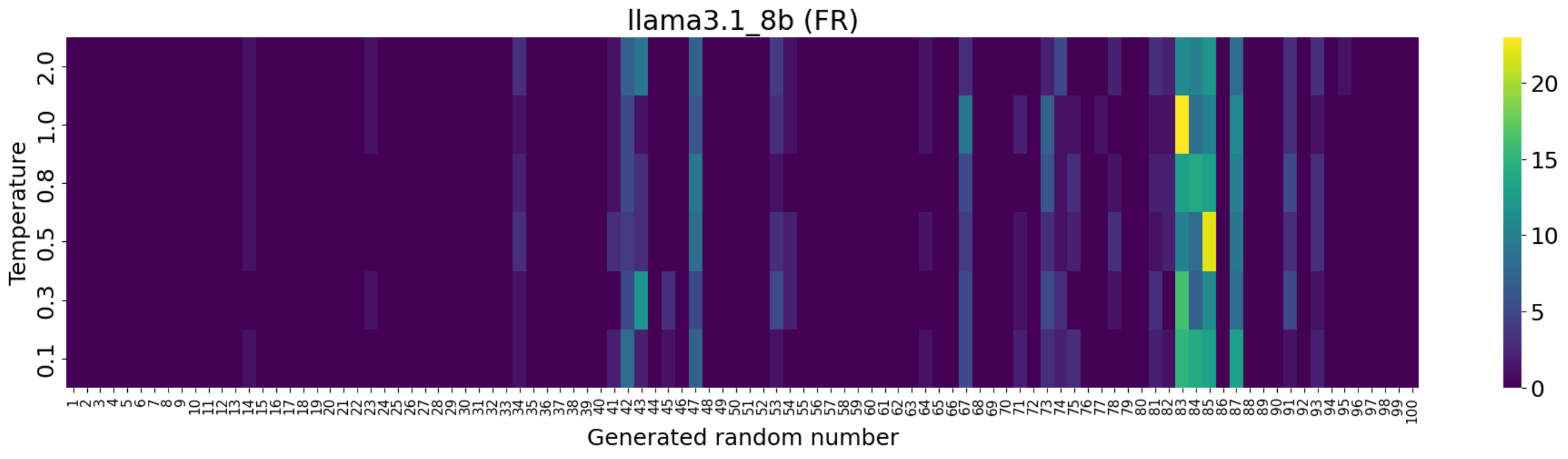}
    \includegraphics[width=1.05\linewidth]{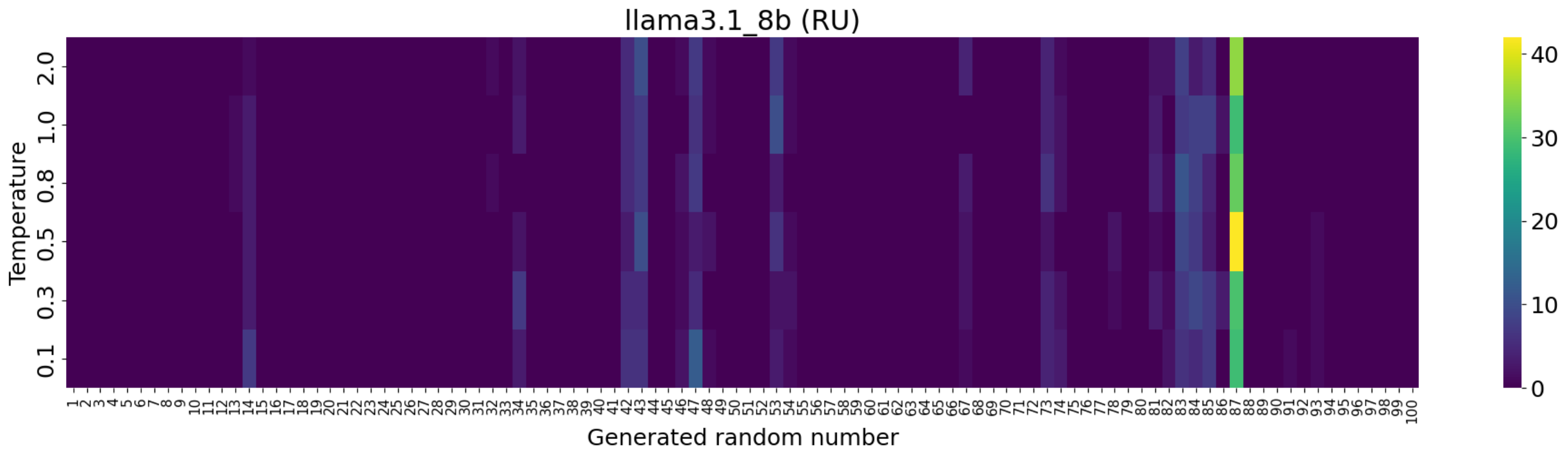}
    \caption{Comparison between four different languages for the generated number distributions in Llama 3.1-8b model in the 1--100 range.}
    \label{fig:barcodes_1_100_llama}
\end{figure}

We can also study the linguistic variance for a single model, as done in Figure \ref{fig:gemini_range_1_5} for the 1--5 range and Gemini 2.0. In this case, we show the results for Llama 3.1-8b in four different languages (Chinese, English, French and Russian) in Figure \ref{fig:barcodes_1_100_llama}.

There are interesting differences between these languages, even for the same LLM. Although Llama 3.1-8b seems to have a preference for numbers in the 42--47 and 81--87 ranges, Chinese and French present more variability than English or Russian. The aversion for upper extreme values is avoided in Chinese and French, which generate numbers over 87 (something not happening in English and Russian). There is no strong dependence of the results with the temperature of the model. These variances across languages for the same model, yet maintaining some of its ``fingerprint'' values, point towards a dual generation bias: on one hand, there is a deeply inherited bias from the training corpus, leading to these systematically repeated values. But on the other hand, there is some uniqueness associated to different languages, suggesting that part of the generation process is affected by the computed values of the self-attention layers depending on the detected language. 

For the 1--100 range the randomness index is less representative in LLMs, as they have the same number of observations that allowed values (100). While in Python we are not restricted and we can generate runs with very large volume of samples, in LLMs we are limited by computational resources. Instead, in this case we present (Figure \ref{fig:violinplots_1_100}) a set of violin plot panels showing the distribution of the different models for four languages in extreme temperatures, as well as a random Python \texttt{randint()} simulation for comparison:

\begin{figure}[!ht]
    \centering
    \includegraphics[width=0.46\linewidth]{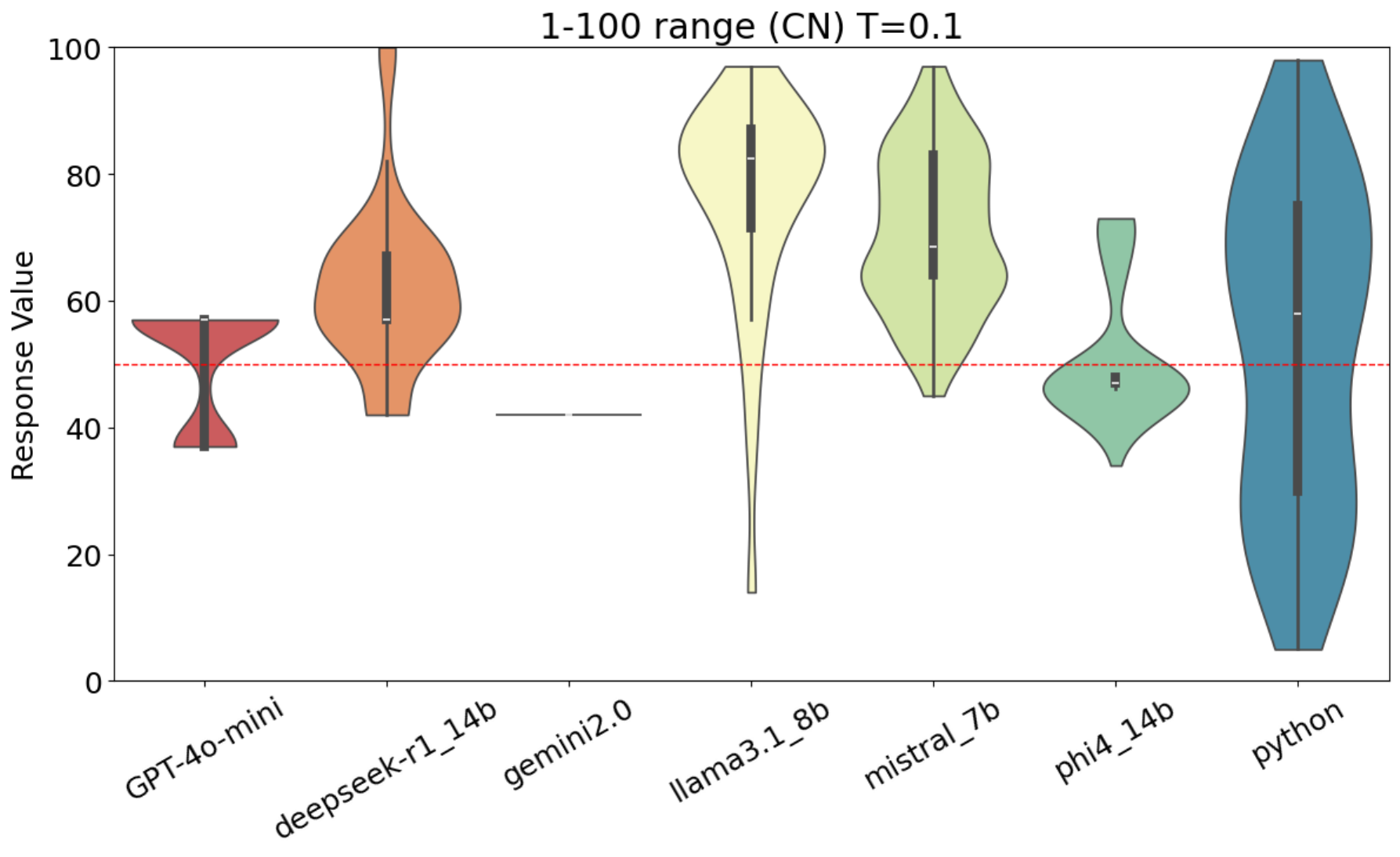}
    \includegraphics[width=0.46\linewidth]{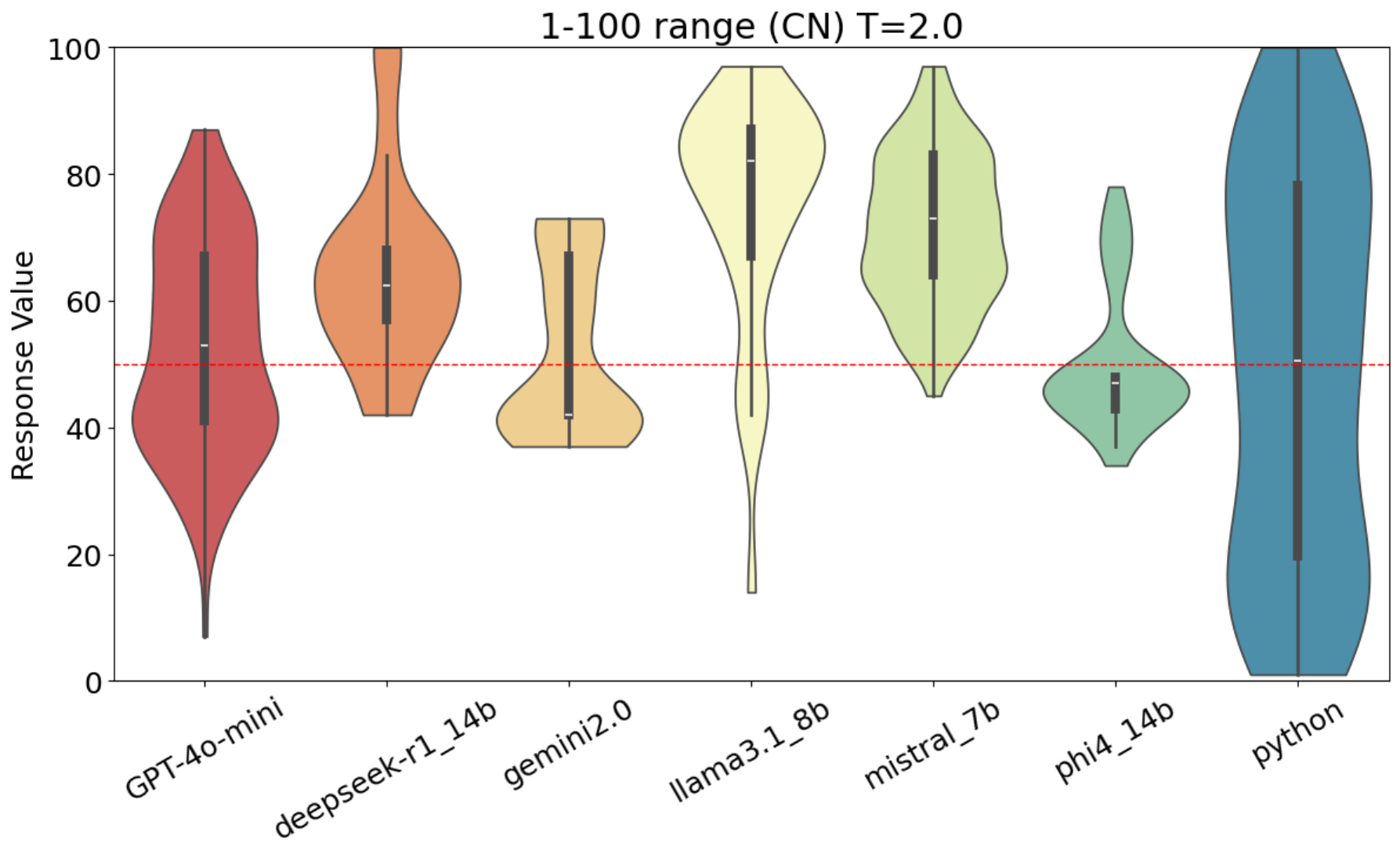}
    \includegraphics[width=0.46\linewidth]{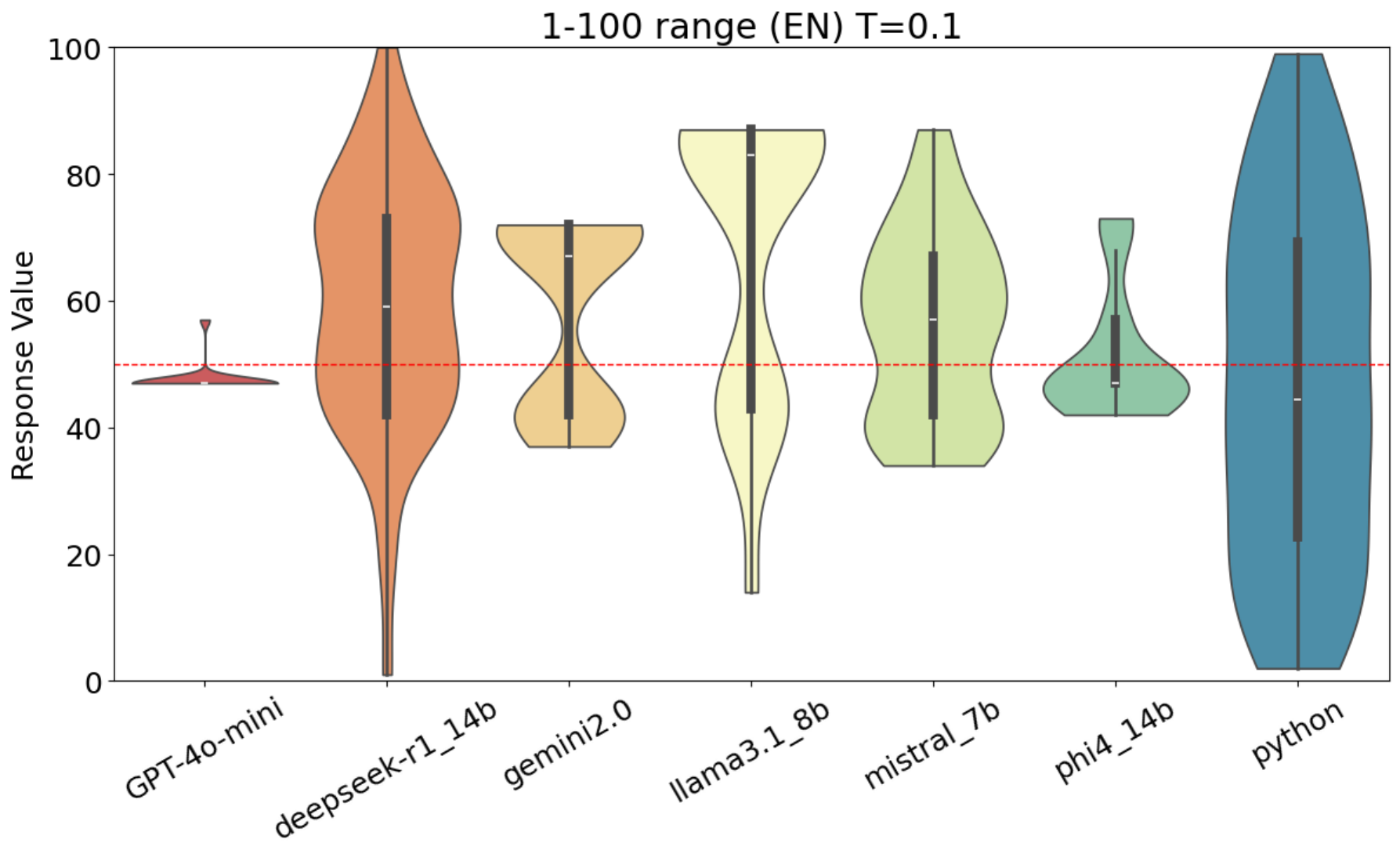}
    \includegraphics[width=0.46\linewidth]{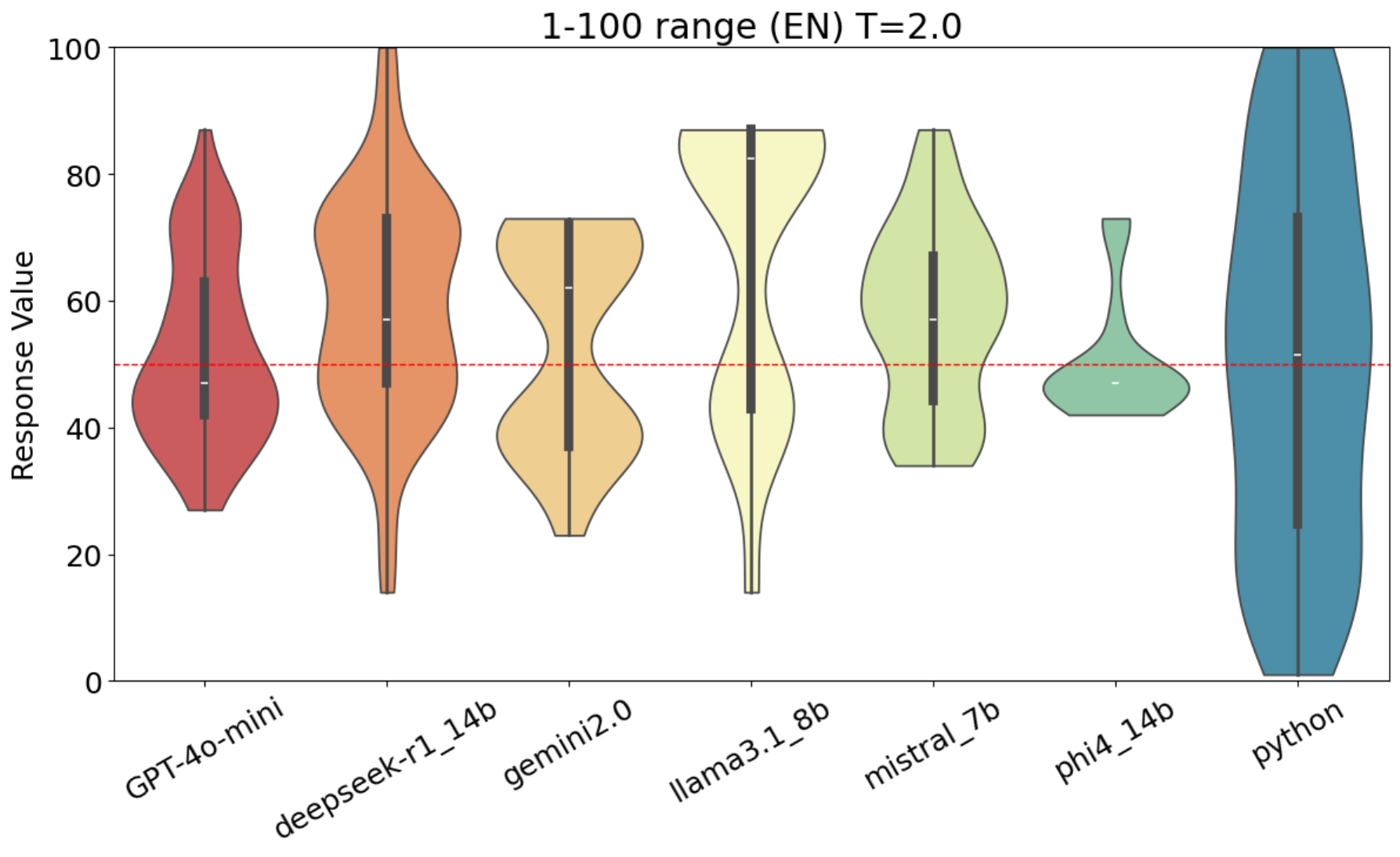}
    \includegraphics[width=0.46\linewidth]{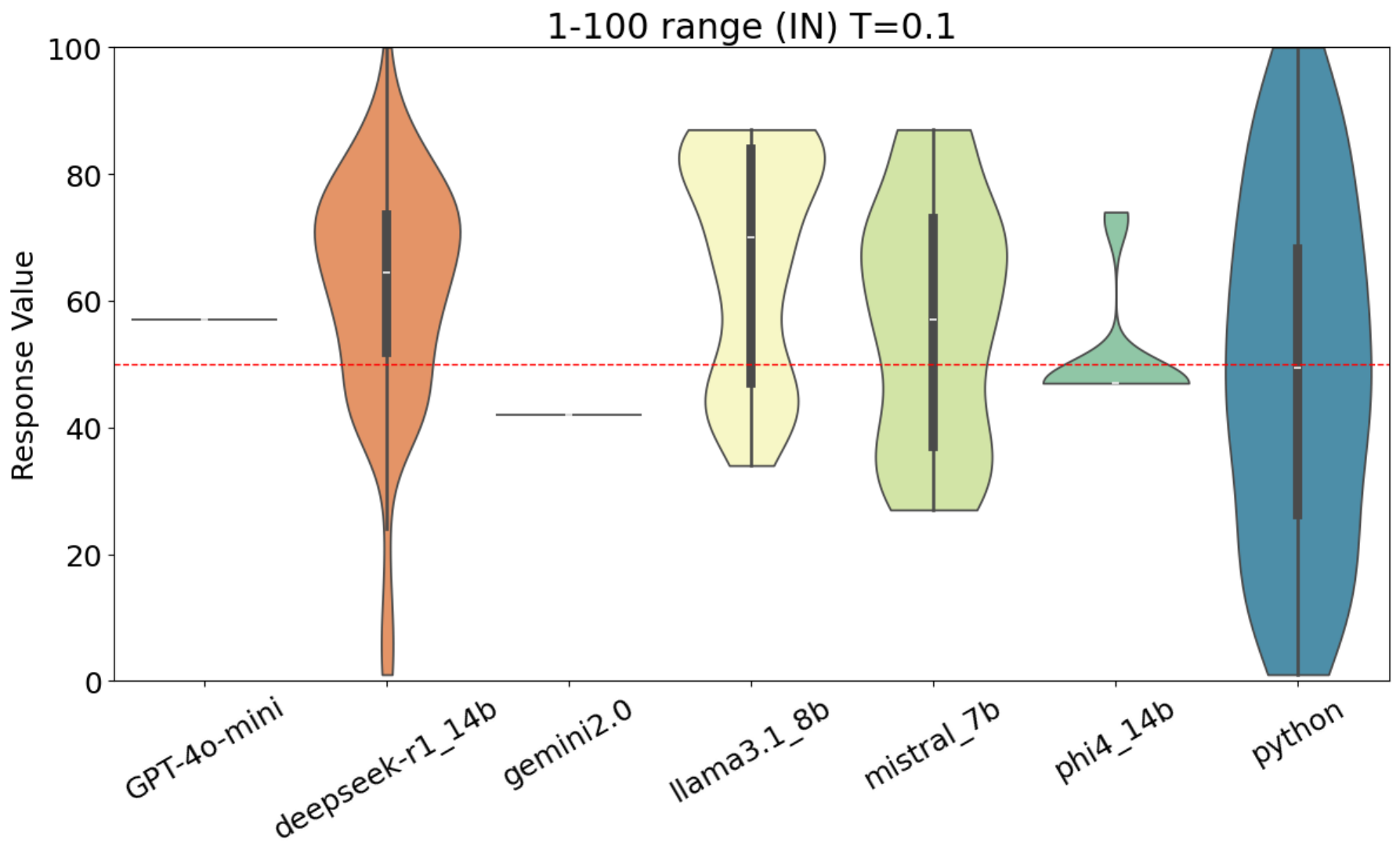}
    \includegraphics[width=0.46\linewidth]{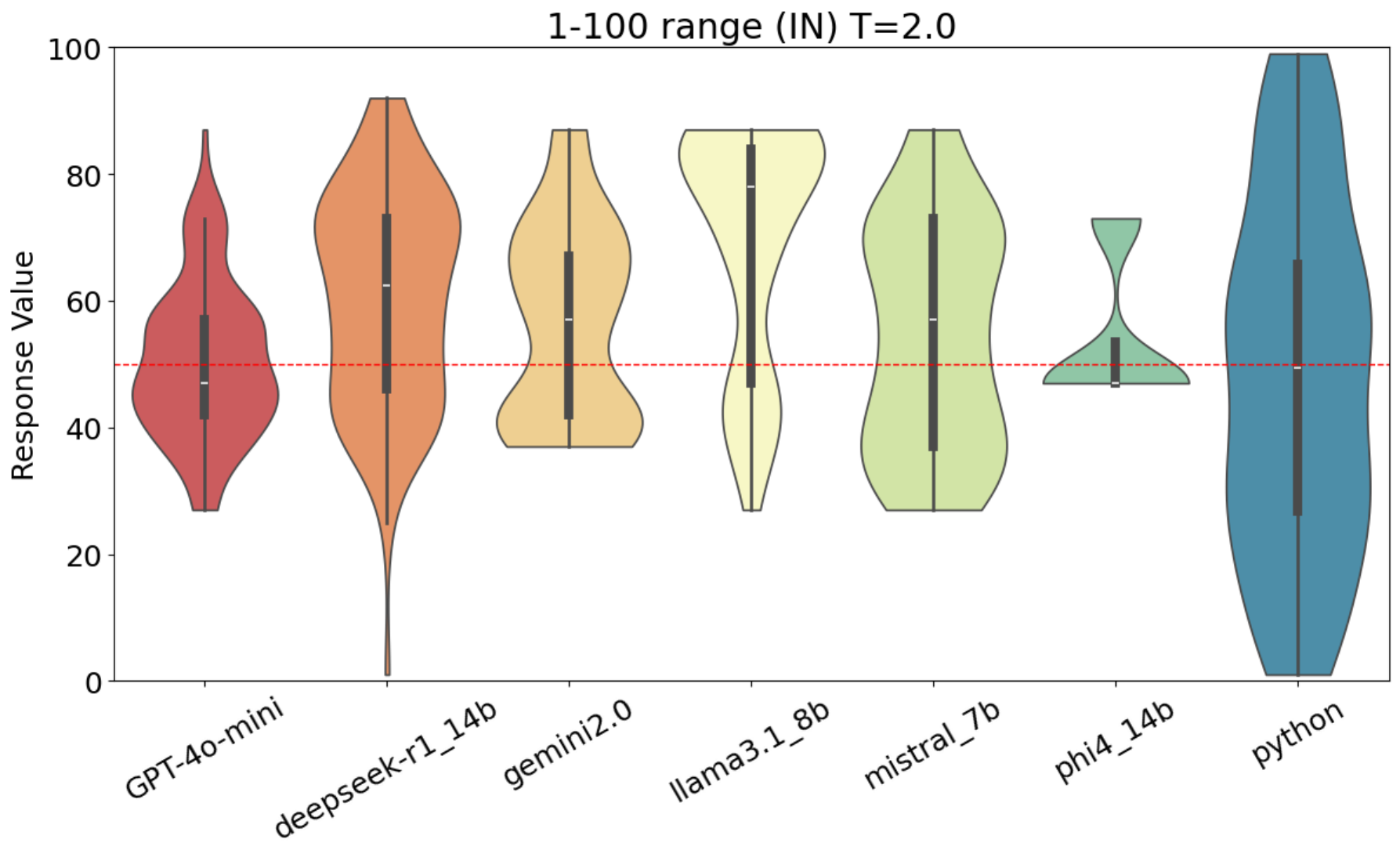}
    \includegraphics[width=0.46\linewidth]{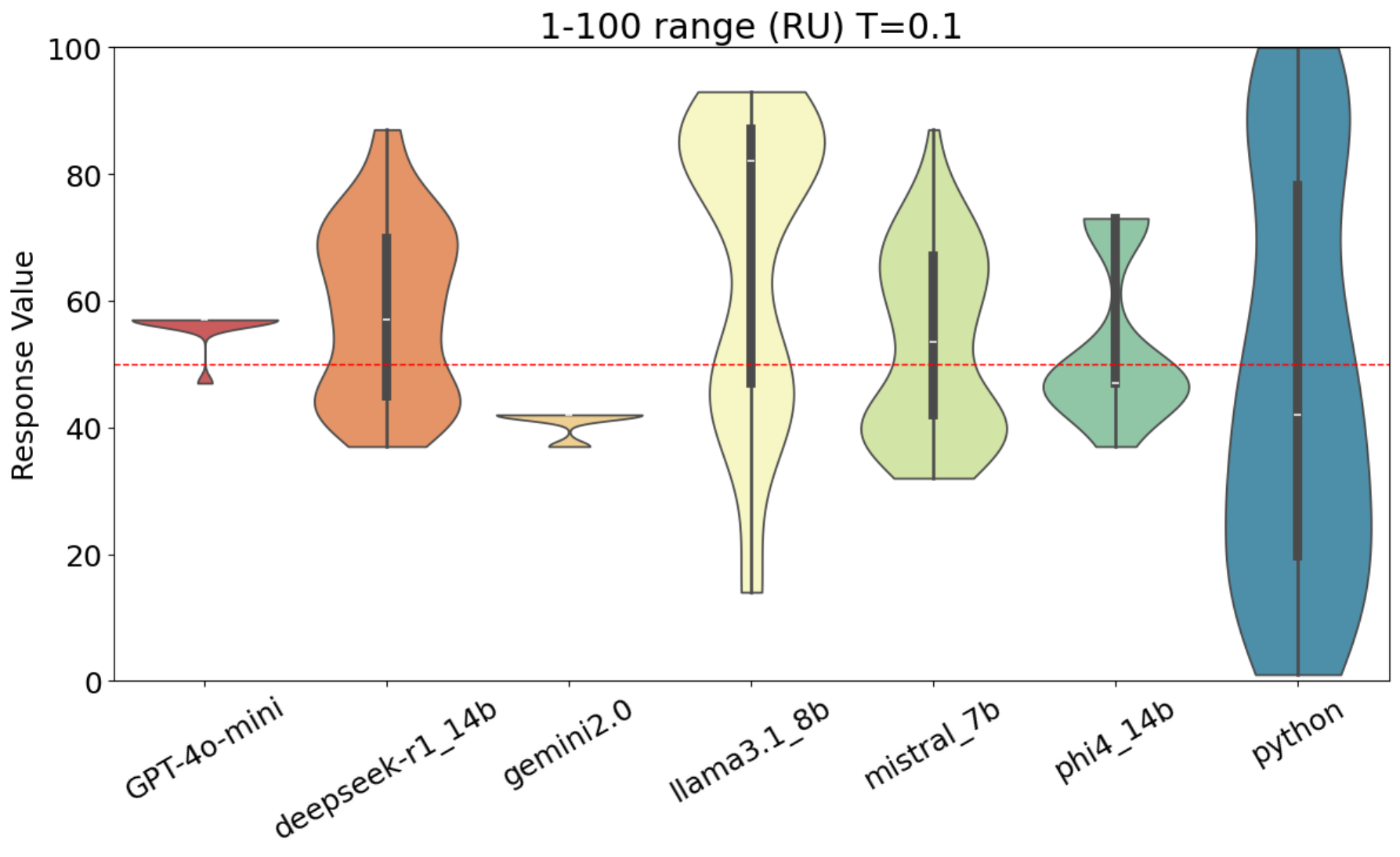}
    \includegraphics[width=0.46\linewidth]{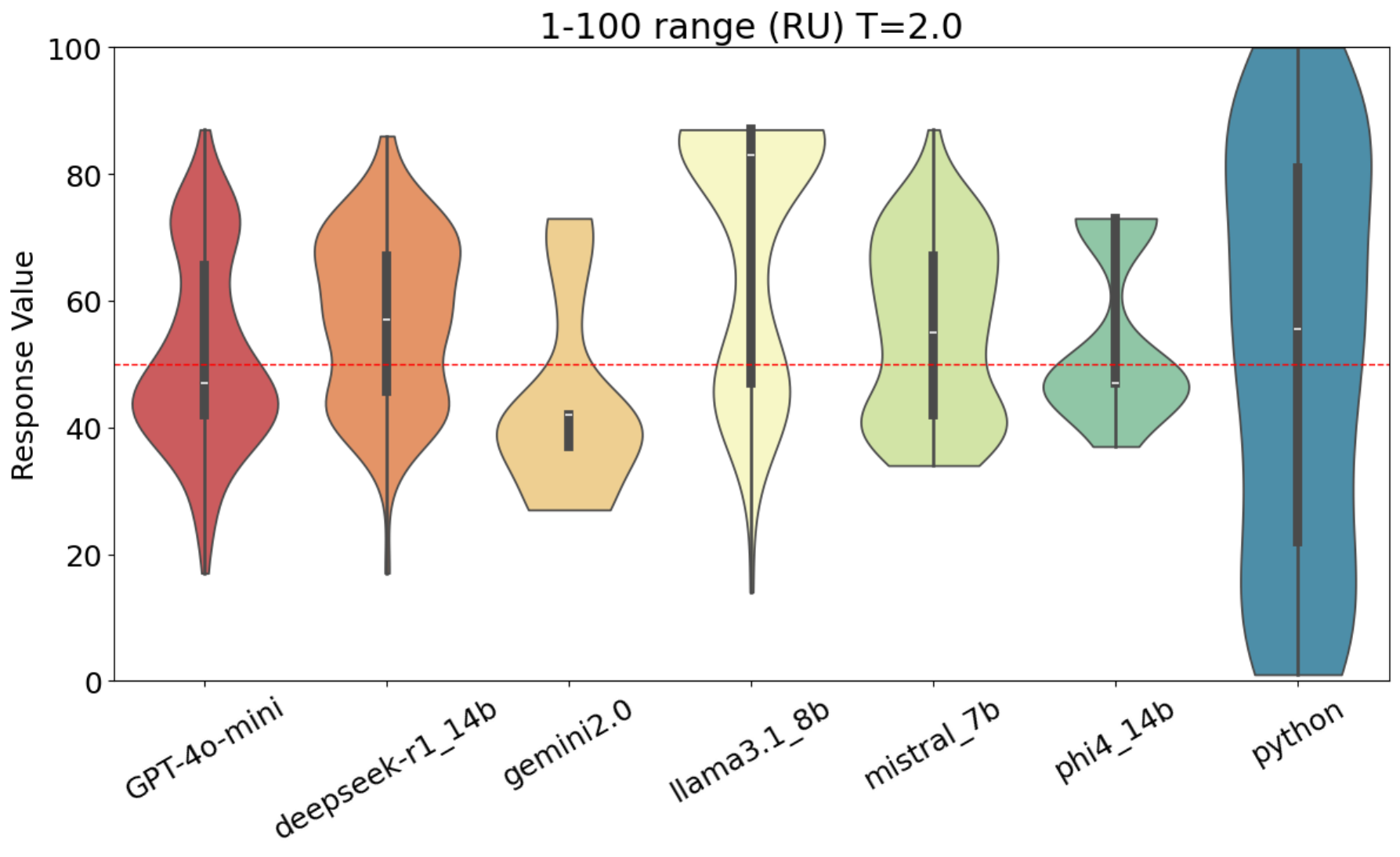}
    \caption{Violin plots for the 1--100 range. Left and right columns show extreme temperatures $(T=0.1, 2.0)$, while rows display different models. Each subpanel features the distribution of generated numbers, with LLMs on the X axis. Additionally, we show for comparison random runs of the Python \texttt{randint} simulations. A horizontal, red dashed line is shown at 50, the central value of the 1--100 range.}
    \label{fig:violinplots_1_100}
\end{figure}

Most models are systematically skewed towards larger values (the dashed, red line shows the middle value of the range: 50), and present less variability than the \texttt{randint} module, even if for all of them only 100 samples are taken, which, as mentioned before, are not enough for a uniform sample. Yet, the Python randomly-selected files all present (within reasonable deviations) the expected distribution, reaching out both to small and large numbers, and with an average close to the range middle value.

Gemini 2.0 and GPT-4o-mini are very limited in this range for $T=0.1$, with extremely narrow violin plots, as much as a line (when there is only one found value), pointing towards a very strong bias in the generation process. It is interesting to note, though, that increasing the temperature help these models, while there is no significant change in other LLMs such as Mistral-7b or Llama 3.1-8b (as already discussed with Figure \ref{fig:barcodes_1_100_llama}). This points towards systematic differences in the training or next token generation process between such models. Specifically, we remind the reader both Gemini 2-0 and GPT-4o-mini are private, API-only accessible models, which may have additional instructions when generating an answer for very low or very high temperatures.

\section{Conclusions}
\label{sec:conclusions}
In this paper, we have studied the biases and determinism of Large Language Models when prompted to generate a random number within a given range. We defined a experimental setup comprising three different ranges (1--5, 1--10, and 1--100), six models (DeepSeek--R1-14b, Gemini 2.0, GPT-4o-mini, Llama 3.1-8b, Mistral-7b, and Phi4-14b), seven different languages (Chinese, English, French, Hindi, Japanese, Russian, and Spanish), and six temperatures (0.1, 0.3, 0.5, 0.8, 1.0, 2.0), comprising a total of 75600 individual calls.

The tested models are heterogeneous and representative of different paradigms, such as nationalities, architectures, number of parameters and access (local vs. API). Large models, such as GPT and Gemini are often regarded as more imaginative and creative; nevertheless, we found that these are as deterministic and biased as their smaller competitors, if not more.

We defined a randomness index (Eq. \ref{eq:ri}) to take into account the range of observed values in relation to the possible values within the range, the standard deviation and the temperature of the model, also including the Shannon entropy. By comparing this index to hundreds of Python \texttt{randint} simulations, we defined objective criteria to quantify how stochastic are the LLM results.

We studied in detail the internal process of DeepSeek--R1-14b, as a reasoning model which outputs a \texttt{<think>} block with a Chain-of-Thought, step-by-step justification of its final answer. Yet, this model did not present significant differences when studying its randomness indices, regardless of the specific configuration.

The prompt language differences, studied in this work for the first time, can shed some light on the internal training and generation processes of these models. In particular, we found some models are systematically less diverse for some languages. DeepSeek--R1-14b internal reasoning is done in Chinese (we remind DeepSeek is a Chinese developer), English or Spanish, while in other languages, most of the times it is done in English, yet sometimes it is done in the prompt language. This suggests these three languages comprise the majority of DeepSeek's training corpus.

We show in Tables \ref{tab:index_range5} and \ref{tab:index_range10} the aggregated results for the randomness index in the 1--5 and 1--10 ranges, computed as average values across all temperatures. We report the average values for each model (across all languages) and for each language (across all models), to study systematic biases. As seen in the table, the most diverse (or less biased) language is Japanese in both ranges, partially helped by the good performance of DeepSeek--R1 in such language. Likewise, the most stochastic model is DeepSeek--R1 in both ranges, yet in the 1--10 is matched by Mistral. The values in the 1--10 range are in general smaller than in the 1--5 range, as there are 10 available values, yet most models only select 2 or 3 values, more penalized by out defined randomness index (see Eq. \ref{eq:ri}) than selecting those 2 or 3 values out of 5 available numbers.

\begin{table}[ht]
    \centering
    \begin{tabular}{|l|ccccccc||c|}
        \hline
        \textbf{Model} & \textbf{CN} & \textbf{EN} & \textbf{ES} & \textbf{FR} & \textbf{IN} & \textbf{JP} & \textbf{RU} & \textbf{Model avg} \\
        \hline
        \hline
        DeepSeek--R1 & 0.06   & 0.05   & 0.05   & 0.05   & 0.04   & 0.16   & 0.02   & \bf0.06   \\
        Gemini 2.0      & 0.02   & 0.01   & 0.02   & 0.02   & 0.002  & 0.009  & 0.01   & \bf0.01   \\
        GPT-4o-mini    & 0.007  & 0.003  & 0.005  & 0.004  & 0.005  & 0.002  & 0.003  & \bf0.004  \\
        Llama 3.1-8b   & 0.06   & 0.02   & 0.08   & 0.02   & 0.02   & 0.08   & 0.02   & \bf0.05   \\
        Mistral    & 0.06   & 0.07   & 0.05   & 0.06   & 0.04   & 0.003  & 0.06   & \bf0.05   \\
        Phi4      & 0.02   & 0.01   & 0.02   & 0.02   & 0.02   & -- & 0.006  & \bf0.02   \\
        \hline
        \hline
        \textbf{Language avg} & \bf0.04   & \bf0.03   & \bf0.04   & \bf0.03   & \bf0.02   & \bf0.05   & \bf0.02   & \bf0.03   \\
        \hline
    \end{tabular}
    \caption{Results of the randomness index for the 1--5 range, with the average computed per model (across all languages) and per language (across all models). Individual values are averaged across the different temperatures.}
    \label{tab:index_range5}
\end{table}

\begin{table}[ht]
    \centering
    \begin{tabular}{|l|ccccccc||c|}
        \hline
        \textbf{Model} & \textbf{CN} & \textbf{EN} & \textbf{ES} & \textbf{FR} & \textbf{IN} & \textbf{JP} & \textbf{RU} & \textbf{Model avg} \\
        \hline
        DeepSeek--R1 & 0.04   & 0.04   & 0.03   & 0.04   & 0.06 & 0.09   & 0.03   & \bf0.05   \\
        Gemini 2.0      & 0.005  & 0.000  & 0.01   & 0.008  & 0.000  & 0.02   & 0.001  & \bf0.007  \\
        GPT-4o-mini    & 0.005  & 0.002  & 0.001  & 0.000  & 0.002  & 0.004  & 0.001  & \bf0.002  \\
        Llama 3.1-8b   & 0.01   & 0.002  & 0.03   & 0.02   & 0.009  & 0.05   & 0.01   & \bf0.02   \\
        Mistral    & 0.06   & 0.02   & 0.02   & 0.03   & 0.11   & 0.03   & 0.03   & \bf0.04   \\
        Phi-4      & 0.001  & 0.000  & 0.002  & 0.000  & 0.000  & -- & 0.000  & \bf0.001  \\
        \hline
        \textbf{Language avg} & \bf0.02   & \bf0.01   & \bf0.02   & \bf0.02   & \bf0.03   & \bf0.04   & \bf0.01   & \bf0.02   \\
        \hline
    \end{tabular}
    \caption{Same as Table \ref{tab:index_range5} but for the 1--10 range.}
    \label{tab:index_range10}
\end{table}

There are several psychological studies on people's choices when prompted the same question. For a low-range of allowed values, like 1--5, people tend to choose the central value 3 or 4, reproducing most of our results \cite{towse2014not}. This is known in psychology as the ``central tendency bias'', or ``risk aversion'' \cite{kahneman2011thinking} which leads us to favor options perceived as average. Likewise, prime numbers are perceived as ``more random'', as they resist simple categorization. In particular, we observe the most popular choices for the different ranges (3 and 4 for 1--5, 5 and 7 for 1--10 and 37, 47, 73 for 1--100) are all prime. The exception is 42, which is a well-known choice for its cultural relevance since mentioned in Douglas Adams' The Hitchhiker's Guide to the Galaxy as the answer to the meaning of life.

Another more informal study\footnote{\href{https://www.reddit.com/r/dataisbeautiful/comments/acow6y/asking_over_8500_students_to_pick_a_random_number/}{Link to the Reddit discussion}} was performed via College Pulse App among ca. 9000 US college students, asking them to choose a number between 1 and 10. The findings here support the central tendency bias, as well as highlighting a strong preference for 7, known as a popular choice for its cultural symbolism.

In the 1--100 range, a 200,000 participants study conducted by YouTube channel Veritaserum\footnote{\href{https://youtu.be/d6iQrh2TK98}{Link to video}} found that people tend to choose numbers containing 7, like 7 itself, 73, 77 and 37. Interestingly, when participants were asked to choose what would be the least-selected number in their opinion, they said 73 and 37, despite the least popular being multiples of 10 (30, 40, 50...), unconsciously perceived as ``too wholesome to be random''. Furthermore, humans are biased towards larger values over lower ones. We did not find any study regarding this phenomenon, yet this is systematically reproduced by our results in the three probed ranges.

In another informal study\footnote{\url{https://llmrandom.straive.app/}}, authors test three different LLMs for the 1--100 range for an English prompt, finding strong biases. While they test previous versions of Gemini (1.0) and GPT (3.5 Turbo), their results are very similar to ours. In the case of GPT-3.5 Turbo, it shows preference for numbers 47 and 57, followed by 42 and 73, coinciding exactly with our results with GPT-4o-mini. For Gemini 1.0 the general results are also identical. This suggests that, although newer versions of the models may update their parameters by including new data (mostly via reinforcement learning from human feedback), the inherent bias remains the same.

Attending to our results, these patterns are replicated by the LLMs, but the output is not being generated from simple occurrences of numbers in training data. If this was the case, according to Benford's law \cite{Wang_2024}, ``1" would be the deterministic choice, especially in the low-range configuration where only 5 values are available. Nevertheless, ``1'' is never selected by LLMs\footnote{Except in the already mentioned cases of Gemini 2.0 and DeepSeek--R1 with Japanese prompting}.

This is explained because the self-attention mechanism is not looking for a simple frequency pattern (such as TF-IDF), but rather understanding the context in which a number appears (in this case, human stochastic choices). LLMs are decoder-only, autorregressive models, where the next token is selected from a pool with probabilities computed from the self-attention mechanism logits of the previous sequence. Thus, it should be expected that in a range 1--X, the probabilities of the tokens corresponding to different numbers in such range were assigned similar values (e.g., $\sim20\%$ for a 1--5 range) by the model. Nevertheless, we observe a probability distribution with clear preferences. Even for very high temperatures, where such distribution should smoothen, the effect is still quite significant, not generating in any case random numbers (and, in particular, avoiding certain values).

Therefore, as LLMs are just predicting a next probable token, they are propagating human preferences, which are themselves a product of cognition biases, such as the aversion to extreme values. This yields an interesting paradox: LLMs are, by construction, non-deterministic. But when asked to behave as such with numbers, they become quite the opposite.

Future work could include the modification of top-p and top-k parameters, which in this study were left them to their default values. Also, a detailed prompt engineering could partially change the results. Giving a range not starting in 1, or a range so big the numbers are composed by more than one token, would also be possible extensions of this line of research to understand systematical bias of LLMs.


\acks{The author acknowledges I.Z. the original idea for this paper. This work has received no funding, and there are no competing interests to declare.}


\newpage

\appendix
\section{Full output of DeepSeek-R1 reasoning process}
\label{app:1}

We reproduce a full output of DeepSeek-R1's CoT reasoning to generate a random number between 1 and 100, with a Spanish prompt and $T=0.8$. During this process more than 30 different numbers are proposed. Also, in the beginning the model claims to have been provided with an example (``57''), which is one of the most popular choices by the tested LLMs. Yet, as explained in Section \ref{sec:methodology}, the prompt does not provide any further instructions or examples.

\begin{tcolorbox}[
  breakable,
  colframe=black,
  colback=white,
  title={DeepSeek full CoT reasoning example}
]
\ttfamily
Okay, so I need to figure out how to respond when the user asks for a random number between 1 and 100 without any extra text. They've already provided an example where they responded with just ""57."" Now, I'm supposed to come up with my own response following that format.
\\
First, I should understand the task clearly. The user wants a single number, no words or explanations—just the number itself. That means when I generate the response, it should be plain text without any additional information or formatting like bold or italics.
\\
I think about how to generate a random number in this range. Since I'm not coding right now, I'll have to do it manually. Maybe I can use a method similar to rolling dice, but since it's just for one instance, I don't need a complex algorithm. Alternatively, I could think of a way to randomly pick a number between 1 and 100 without any bias.

Let me try to visualize it. If I imagine a list from 1 to 100 and pick one at random, what would that number be? Sometimes people use the current second or other time-based methods, but since I'm just doing this mentally, I can think of it as a simple selection.
\\
Wait, maybe I can use an online random number generator if I could access it, but in this case, I have to do it manually. So perhaps I'll count on some method like using the digits from a book or something. Alternatively, since I'm just creating a single response, I can decide arbitrarily as long as it's within 1-100.
\\
So, thinking of numbers... maybe 73? No, that seems too high. Or wait, perhaps I should go lower. Let's see, 24? That could work. But then again, why not a higher number like 89? Hmm, this is tricky without any method.
\\
Alternatively, I can think of the time. If it were 3 PM when I start thinking about this, maybe take the minutes and seconds or something. Wait, that's not applicable here since I'm just in my mind. Maybe I'll settle on a number like 42. It's a common answer for such questions, but perhaps the user wants a genuine random one.
\\
Wait, no, I should make sure it's truly random. Let me try to think differently. Imagine flipping a coin multiple times or using some other method to generate entropy, but since that's complicated without tools, maybe just choosing based on the letters in my name or something else.
\\
Alternatively, perhaps I can use the last two digits of a phone number or something like that. But again, without external data, it's hard. Maybe I'll go with 53. It seems arbitrary enough and falls within the range.
\\
Wait, but I have to make sure it's between 1 and 100, inclusive. So numbers like 100 are possible as well. Let me think of a number closer to the middle, say around 50. Maybe 67? That seems okay.
\\
Alternatively, since I need to be precise, perhaps I can visualize a spinner or something. Imagine a wheel divided into 100 equal parts and randomly stopping on one. It could land anywhere, so let's just pick a number that's not too common but also not too high or low.
\\
Wait, maybe using the number of letters in a specific word. For example, the word ""random"" has 6 letters. But that might not be random enough. Alternatively, take the sum of digits from a specific date, like today's date. If today is the 15th day of the month, and the year is 2023, then maybe 1+5=6 or something, but that's too low.

Alternatively, think of the last two digits of a famous number, like pi (3.14159...), but again, this might not be truly random.

Perhaps I can use a method where I count on my fingers to generate a seed for randomness. Let me try that. Starting from 1 and counting up to 100, landing randomly somewhere—maybe around the middle is better. So, if I land on the 50th number, it would be 50. But since I'm just imagining, perhaps 74? That's in the higher half.

Wait, maybe I should think of a way to average it out. If I split 1-100 into two halves, lower and upper, then pick randomly between them. Let's say heads for lower (1-50) and tails for upper (51-100). But without flipping a coin, how do I decide? Maybe using the initial thought process to decide.

Alternatively, maybe think of it as a percentage: 25\% chance for each quarter. So, if I randomly pick a quarter—first 25 numbers, next 25, etc.—but since I can't really simulate that mentally, perhaps just selecting a number in my mind.

Wait, another approach: take the current second or minute and use modulo 100 to get the number. For example, if the current time is 3:47 PM, then 47 seconds—so 47 mod 100 is 47. That could work. But since I'm just thinking, I don't have a real-time reference.

Alternatively, think of my birth year's last two digits as a base and add some random number to it within the range. For example, if someone was born in 1985, take 85 and then adjust by adding or subtracting numbers until it lands between 1-100. But that might be too methodical.

Wait, perhaps I can use the page number of a book I'm reading. If I'm on page 73, then 73 is my number. Alternatively, if I think of a favorite song's track number in an album—say, track 15—and use that. But again, without external data, it's hard.

Maybe the key here is to just pick a number I feel like, as long as it's within the range and doesn't have any significance. Let me try to do that now. Okay, closing my eyes and thinking of a number... 34? That seems okay.

Wait, but maybe too low. Alternatively, think of a number that's often considered lucky, like 7 or 13. But since the user wants a random one, perhaps not those. Maybe 92 is better—it's in the higher range and doesn't have common associations.

Alternatively, thinking of my shoe size—oh, wait, mine is 8, but that's too low again. Hmm.

Wait, maybe using the number of keys on a piano—which has 88 keys—so taking 88 mod 100 gives 88, but that might be predictable.

Alternatively, think of the atomic number of an element—if I take, say, gold which is 79, but again, it's a known fact. Maybe better to avoid elements with high numbers like that.

Wait, perhaps using my lucky lottery ticket numbers from last week: 23 and 45. But since I need just one, maybe 34? Or 17?

Alternatively, think of the number of letters in a sentence. For example, ""The quick brown fox jumps over the lazy dog"" has 43 letters—so 43 could be the number.

But that might not be purely random as it's based on text length. Alternatively, maybe I can count the number of stars I see when closing my eyes—if I imagine seeing 17 stars, then 17 is the number.

Alternatively, think of a time in minutes since midnight—say, if it's 2:30 PM, that's 90 minutes past noon. But without a real reference, it's hard to use this method.

Wait, perhaps I can just randomly pick a number between 1-100 using a simple thought process. Let me do that now. Okay, starting from 1 and counting up... but that would take too long. Alternatively, think of the digits on a clock: if it's 2:45, then maybe 24 or 45—so 45 is within range.

Alternatively, think of the area code where I live—if it's 555, then 55—but that's too low. Alternatively, take the last two digits of a phone number I know, but again, not random.

Wait, perhaps using the Fibonacci sequence: 1, 1, 2, 3, 5, 8, 13, 21, 34, 55, 89... So if I randomly pick one from there, maybe 34 or 55. But that's not truly random; it's based on a sequence.

Alternatively, think of the number of letters in famous landmarks—like ""Eiffel Tower"" has 12 letters, so 12—but again, too low.

Wait, perhaps using the number of planets in our solar system—which is 8—but that's also too low. Alternatively, the number of elements in the periodic table, which is around 118, but that's over 100.

Hmm, this is getting complicated. Maybe I should just pick a number based on how I'm feeling right now. If I feel energetic, maybe a higher number like 92. If calm, perhaps a lower one like 17.

Alternatively, think of the stock market index numbers—if it's at 34,000 points, then 34 is within range—but that's again too context-dependent.

Wait, maybe I can use a mathematical approach. Let me take the current second in my mental clock. If I estimate that about 5 minutes have passed since I started thinking (which is roughly 300 seconds), so 300 mod 100 is 0. But that's not random.

Alternatively, think of the sum of digits in a number—if I pick a date like 2/29/24, the sum is 2+2+9+2+4=17—so 17 could be my number.

But again, this method isn't purely random. It's based on external factors or calculations.

Wait, perhaps using my heartbeat rate—say, around 60 beats per minute, so in a second, it's about 1 beat every second, which would give me 60 seconds—but that's not helpful here.

Alternatively, think of the number of keys I pressed to type this message—if I've typed around 200 characters, then 20 mod 100 is 20—so 20 could be my number.

But again, it's too context-dependent and might not be truly random.

Maybe the simplest approach here is to just close my eyes and think of a number between 1-100 without any constraints. Let me try that now. Okay, starting from 1... but I don't want to count; I need to visualize a number randomly in my mind. Maybe 47 comes up because it's halfway through the range.

Alternatively, thinking about sports—like basketball players have jersey numbers up to around 98 or so—so maybe 32 is a common number, but that's too known.

Wait, perhaps using initials: if my name is John Doe, and I take J=10 and D=4 in the alphabet, then 10+4=14—but that might not be random enough.

Alternatively, think of the sum of my siblings' ages—if there are three siblings aged 5, 8, and 12, their sum is 25—so 25 could be my number.

But again, this depends on external factors. I think it's time to just make an arbitrary choice without overcomplicating it.

Okay, let me visualize a number between 1-100... Maybe 37? Because I've heard that 37 is considered a lucky number in some cultures—so perhaps 37.

Alternatively, thinking of the number of letters in ""Dame un número aleatorio entre 1 y 100""—that's Spanish for ""Give me a random number between 1 and 100."" Counting the letters: D-A-M-E- space-U-N-O etc. Wait, maybe that's too time-consuming.

Alternatively, think of my favorite song's track number in an album—say, track 15. So, 15 could be my number.

Wait, but I'm overcomplicating again. Maybe it's better to just pick a number without any reasoning, like flipping a coin multiple times to decide each digit—but since I can't flip a coin mentally, perhaps I'll go with an arbitrary choice.

Alright, after all this thinking, maybe 43 is my random number because it's the result of adding 17 and 26 (which are both numbers that came up earlier in different contexts). So, 17+26=43. That seems reasonable.

Wait, but I'm still overcomplicating it. Maybe I should just stop trying to find a method and pick a number without any calculation. Let me try again: 76. Why? Because when I think of the word ""success,"" it's often associated with 76 (as in ""76 times"" you try before succeeding), but that's not really random.

Alternatively, maybe I can think of the number as a combination of digits from my phone number—say, 5 and 2 make 52. So, 52 could be it.

Wait, perhaps using the digits from today's date: if today is 12/03/2023, then adding them up: 1+2+0+3+2+0+2+3=13—so 13 is my number. But that might be too predictable.

Wait, another approach: think of a mathematical constant or famous number like pi (3.1415...) and take the first two digits after the decimal: 14—but that's too low. Alternatively, take more digits for higher numbers.

Alternatively, think of the square root of 100, which is 10—so maybe 10, but that seems too obvious.

Wait, perhaps I can use a mnemonic device or a phrase where letters correspond to numbers. For example, ""A""=1, ""B""=2, etc. But that's time-consuming and might not yield a truly random number.

Alternatively, just pick a number from a song lyric—like ""45 RPM"" which is 45. So maybe 45 could be my number.

But honestly, after all this thinking, I realize that trying to generate a truly random number without any constraints or external aids is challenging. The best approach might be to accept the first number that comes to mind without overthinking it.

So, let me try once more: When I think of ""random,"" sometimes 56 comes up because it's halfway between 50 and 62—but no, that's not random either.

Wait, perhaps using a book's page number where I'm currently reading. If I'm on page 73, then maybe 73 is my number. But again, this depends on external factors.

I think I need to accept that without an actual method or tool, it's hard to generate a truly random number mentally. So, perhaps the best way is to just pick a number and go with it. Let me settle on 42 because it's a well-known answer to life, the universe, and everything—but that might be too cliché.

Alternatively, maybe I can count down from 100: starting at 100, subtracting random increments. But without actual randomness, this won't help much.

Wait, perhaps using binary digits mentally—if I flip a coin five times to generate a number up to 32 (since 5 bits can represent numbers up to 31), but that's too limiting and not truly random.

Alternatively, think of the Fibonacci sequence—numbers like 21, 34, 55—but those are too structured.

Wait, maybe I can use a mathematical function. If I take the current time in seconds since midnight modulo 100, it would give me a number between 0 and 99. But without knowing the exact time, this isn't feasible.

Hmm, perhaps I can just choose 7 because it's considered lucky—so maybe 7 is my number.

Wait, but earlier I thought of 43, then 52, then 45, etc. Maybe after all these thoughts, the most random number that comes up is 43.

I think I'll settle on 43 as my random number between 1 and 100.
\end{tcolorbox}

\section{DeepSeek--R1 outside the prompted range}
\label{app:2}

\begin{figure}[!ht]
    \centering
    \includegraphics[width=0.49\linewidth]{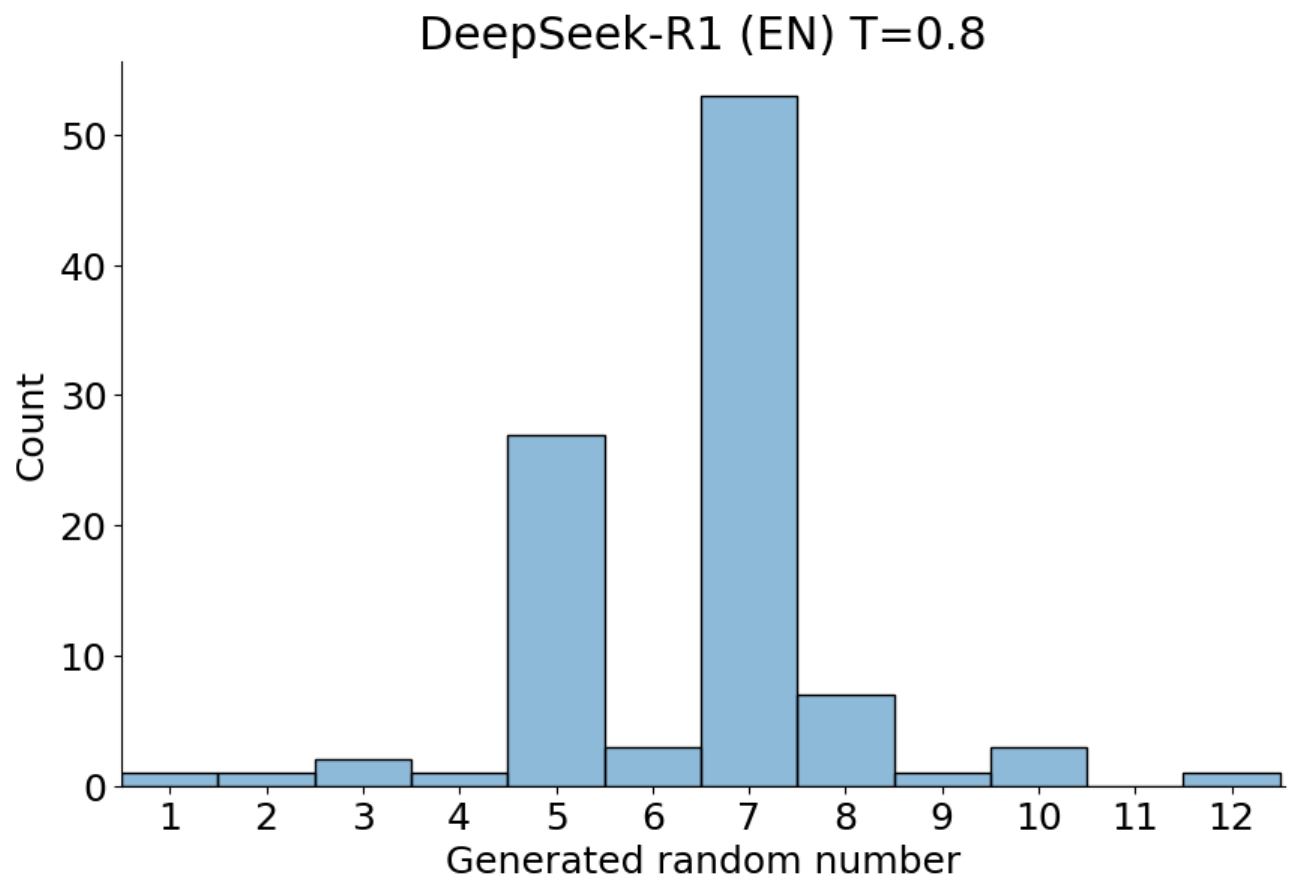}
    \includegraphics[width=0.49\linewidth]{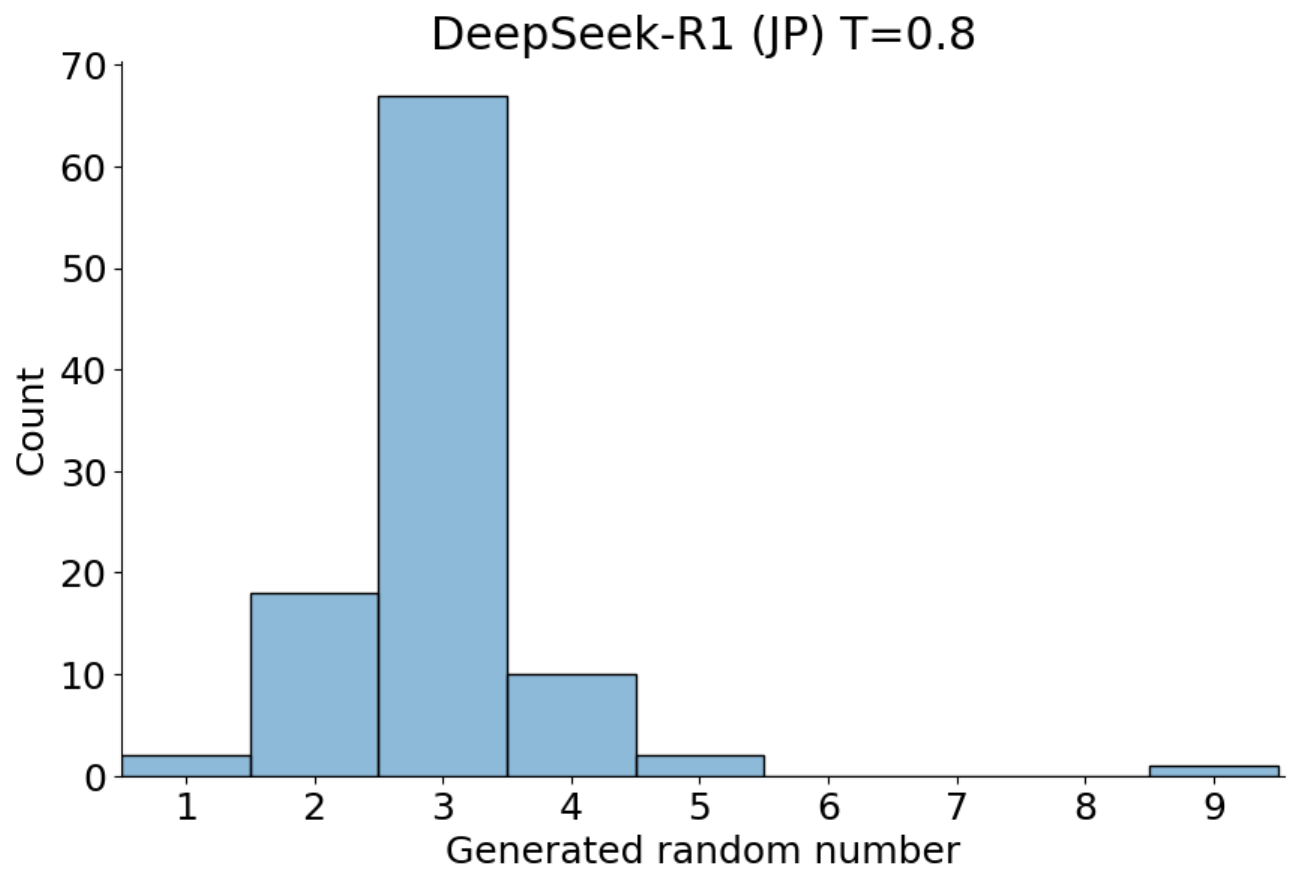}
    \caption{Distribution of the two cases with outside-the-range values. \textbf{Left panel:} DeepSeek--R1 with $T=0.8$, English prompt in the 1--10 range. \textbf{Right panel:} DeepSeek--R1 with $T=0.8$, Japanese prompt in the 1--5 range. Note the results ``12'' and ``9'' respectively, outside the permitted ranges.}
    \label{fig:deepseek_out_of_range}
\end{figure}

\noindent
As already mentioned in Section \ref{sec:results}, in two single cases a model (DeepSeek--R1) chooses a number outside the prompted range. In Figure \ref{fig:deepseek_out_of_range} we show their distributions, as it is interesting to note that in both cases every possible number is selected at least once. The randomness index in these cases is distorted, as there is one extra found value that shouldn't be taken into account. Even so, they both present a low randomness index (0.08 and 0.15), expected from the peaks at ``5'' and ``7'' in the 1--10 range, and ``3'' in the 1--5 range, while every other value is marginally found.

\section{Tests with 1000 calls for the 1--100 range}
\label{app:3}

\noindent
As explained in Section \ref{subsec:high_range}, 100 samples seem not enough when considering a total range of 100 possible values (range 1--100), yet we show LLMs tend to repeat the same few numbers over and over.

\begin{figure}[!ht]
    \centering
    \includegraphics[width=0.99\linewidth]{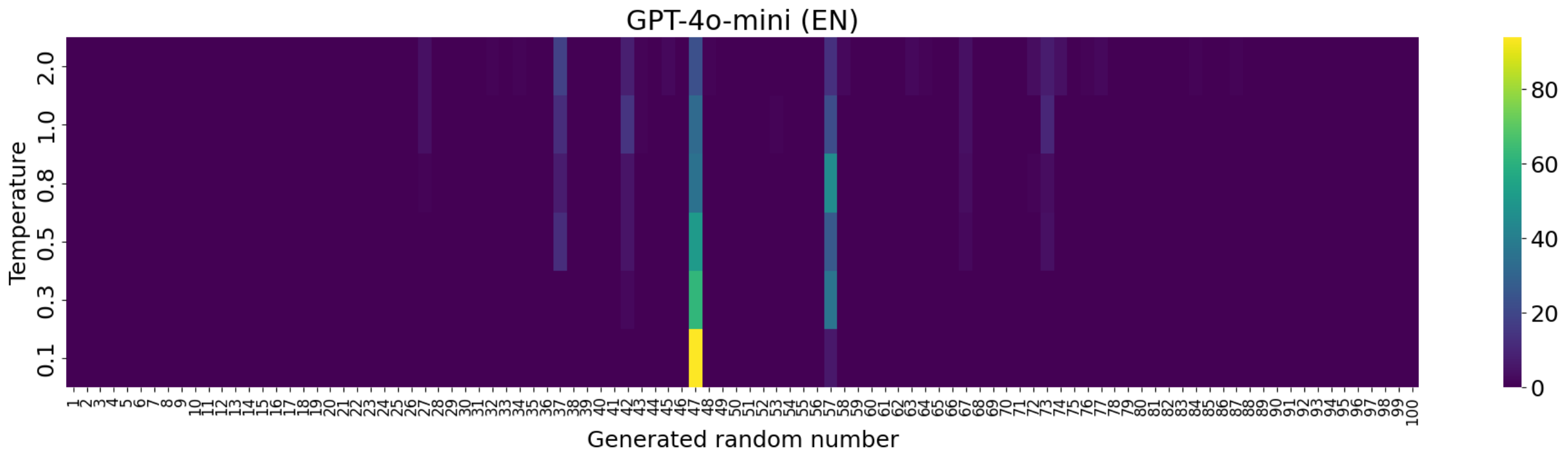}
    \includegraphics[width=0.99\linewidth]{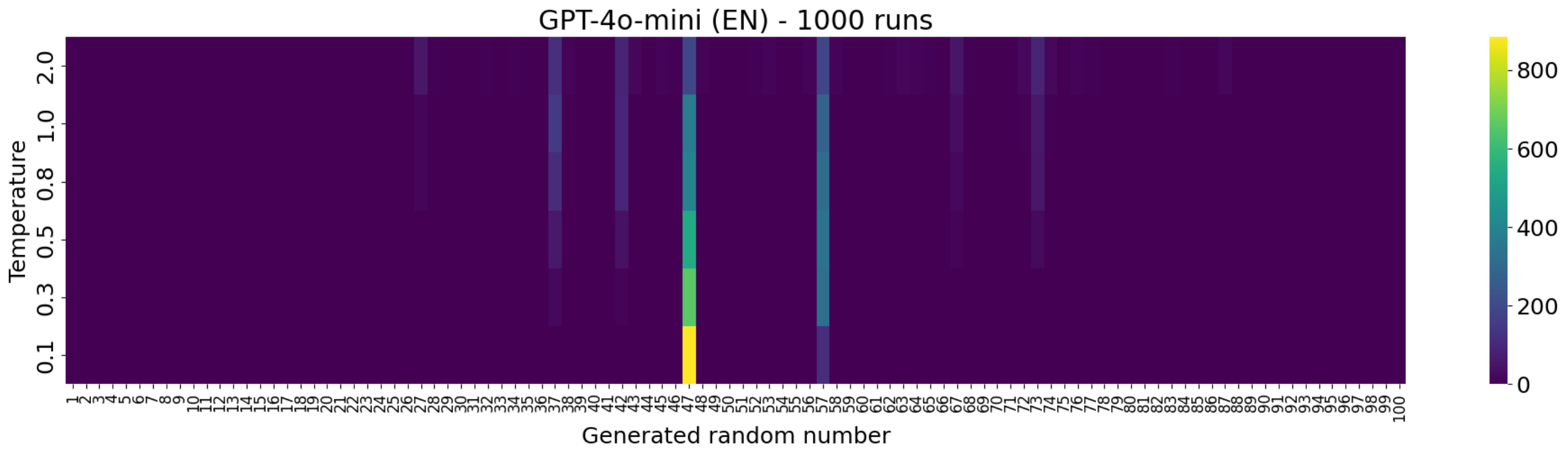}
    \caption{Heatmaps for the 1--100 range configuration showing the distribution of the generated random numbers (X axis) depending on the temperature of the model (Y axis), for the GPT-4o-mini model with English prompt. Upper panel is the standard, 100--call run while lower panel is the 1000 samples test.}
    \label{fig:1000_runs}
\end{figure}

In this Appendix, we show the results of enlarging a factor 10 the sample for the 1--100 range, using GPT-4o-mini in English. We test how representative is a 100-value sample of a 1000-value one, and show the result in Figure \ref{fig:1000_runs}, where we reproduce its 100-calls equivalent.

\begin{figure}[!ht]
    \centering
    \includegraphics[width=0.9\linewidth]{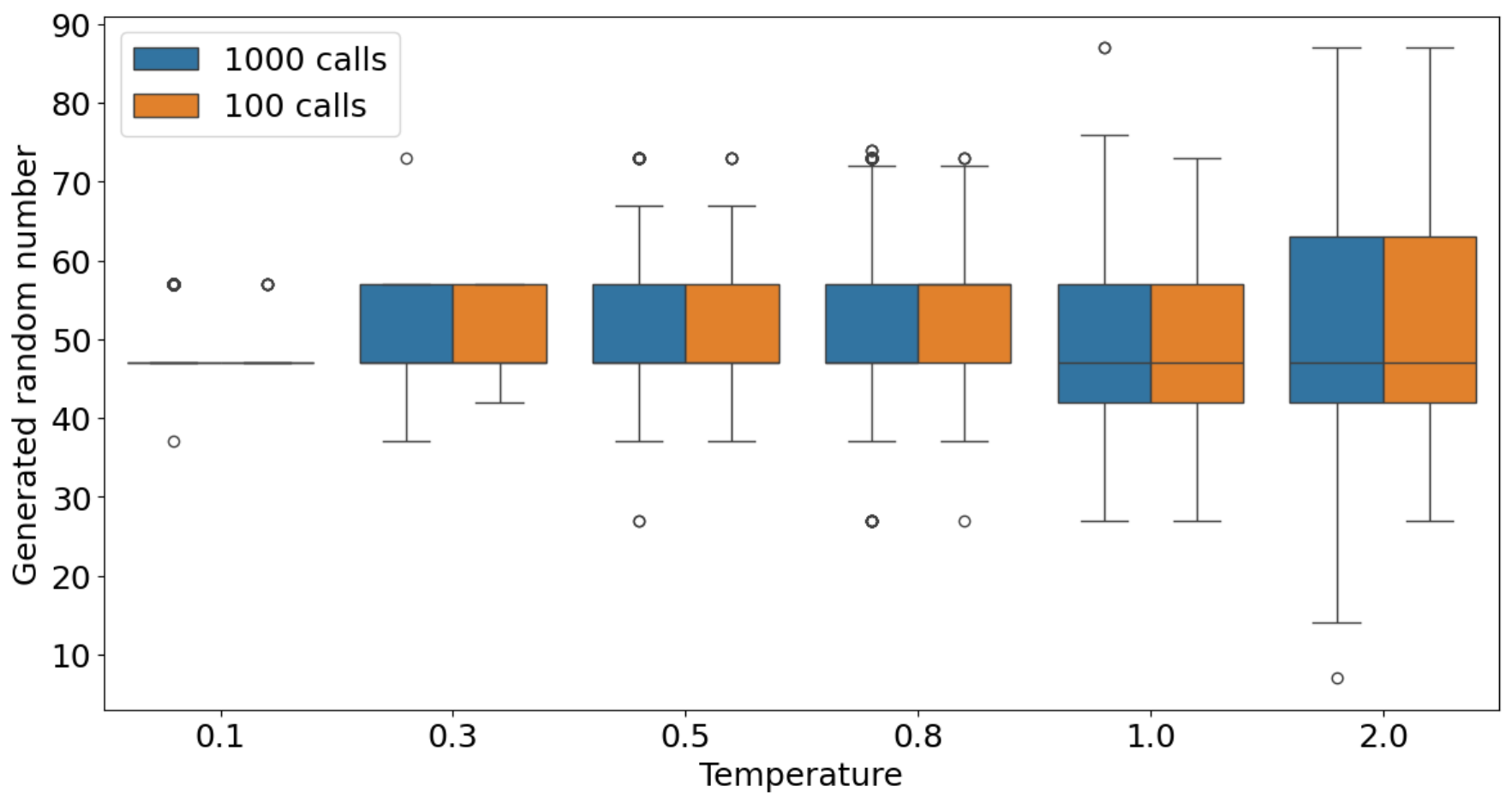}
    \caption{Boxplots for the 1--100 range configuration, for the GPT-4o-mini model with English prompt, showing the 1000-call runs (blue) and standard, 100-call runs (orange). Individual points are outliers computed as those outside 1.5 times the inter-quartile range of the distribution.}
    \label{fig:boxplots_gpt}
\end{figure}

As seen in the Figure, the model is just filling the same numbers, with minimal variations. For example, in the temperature central value $T=0.8$ there are 8 unique values for the 100 sample run, whereas only 2 extra unique values are added in the 1000--call case. Additionally, we show in Figure \ref{fig:boxplots_gpt} the boxplot distribution for both cases. The quartiles and average values are the same for both configurations, showing only some differences for the highest temperatures, where the 1000-call run partially beats the ``phobia'' to low numbers.

\vskip 0.2in
\bibliography{references}

\end{document}